\documentclass[journal]{IEEEtran}
\usepackage{cite}
\usepackage{amsmath,amssymb,amsfonts}
\usepackage{algorithmic}
\usepackage{graphicx}
\usepackage{textcomp}
\usepackage{amsmath,amsfonts}
\usepackage{algorithmic}
\usepackage{algorithm}
\usepackage[algo2e]{algorithm2e}

\usepackage{array}

\usepackage{stfloats}
\usepackage{url}
\usepackage{verbatim}
\usepackage{amsmath}
\usepackage{amssymb}
\usepackage{booktabs}
\usepackage{colortbl}
\usepackage{hhline}
\usepackage{array,multirow,graphicx}
\usepackage{pifont}
\usepackage{setspace}
\usepackage{caption}
\usepackage{comment}
\usepackage{xcolor}
\usepackage{hyperref}
\usepackage{bm}
\usepackage[caption=false,font=normalsize,labelfont=sf,textfont=sf]{subfig}

\usepackage[normalem]{ulem}

\definecolor{darkblue}{rgb}{0.0,0.0,1.0}
\hypersetup{colorlinks,breaklinks,
            linkcolor=darkblue,urlcolor=darkblue,
            anchorcolor=darkblue,citecolor=darkblue}

\ifCLASSINFOpdf
\else
\fi

\begin{document}

%
\title{Hierarchical Semi-Supervised Active Learning for Remote Sensing}
%
%
%

\author{Wei Huang, ~\IEEEmembership{Student Member,~IEEE}, 
Zhitong Xiong,~\IEEEmembership{Member,~IEEE}, Chenying Liu, \\
and Xiao Xiang Zhu,~\IEEEmembership{Fellow,~IEEE}

\thanks{ This project is jointly supported by the German Research Foundation (DFG GZ: ZH 498/18-1; Project number: 519016653), by the German Federal Ministry for the Environment, Nature Conservation, Nuclear Safety and Consumer Protection (BMUV) based on a resolution of the German Bundestag (grant number: 67KI32002B; Acronym: \textit{EKAPEx}), and by the Munich Center for Machine Learning. }

\thanks{Corresponding author: Xiao Xiang Zhu.}
\thanks{
W. Huang, Z. Xiong, C. Liu, and X. Zhu are with the Chair of Data Science in Earth Observation, Technical University of Munich, 80333 Munich, Germany (e-mail: \{w2wei.huang, zhitong.xiong, chenying.liu, xiaoxiang.zhu\}@tum.de).
}
\thanks{C. Liu and X. Zhu are also with the Munich center for Machine Learning, 80333 Munich, Germany.}
}

\markboth{IEEE Transactions on Geoscience and Remote Sensing}%
{Shell \MakeLowercase{\textit{et al.}}: Bare Demo of IEEEtran.cls for Journals}
%


\maketitle

	\begin{abstract}
		The performance of deep learning models in remote sensing (RS) strongly depends on the availability of high-quality labeled data. However, collecting large-scale annotations is costly and time-consuming, while vast amounts of unlabeled imagery remain underutilized. To address this challenge, we propose a \textbf{Hierarchical Semi-Supervised Active Learning (HSSAL)} framework that integrates semi-supervised learning (SSL) and a novel hierarchical active learning (HAL) in a closed iterative loop. In each iteration, SSL refines the model using both labeled data through supervised learning and unlabeled data via weak-to-strong self-training, improving feature representation and uncertainty estimation. Guided by the refined representations and uncertainty cues of unlabeled samples, HAL then conducts sample querying through a progressive clustering strategy, selecting the most informative instances that jointly satisfy the criteria of \textbf{\textit{scalability}}, \textbf{\textit{diversity}}, and \textbf{\textit{uncertainty}}. This hierarchical process ensures both efficiency and representativeness in sample selection. Extensive experiments on three benchmark RS scene classification datasets, including UCM, AID, and NWPU-RESISC45, demonstrate that HSSAL consistently outperforms SSL- or AL-only baselines. Remarkably, with only \textbf{8\%}, \textbf{4\%}, and \textbf{2\%} labeled training data on UCM, AID, and NWPU-RESISC45, respectively, HSSAL achieves over \textbf{95\%} of fully-supervised accuracy, highlighting its superior label efficiency through informativeness exploitation of unlabeled data. Our code will be publicly available.
	\end{abstract}

	
	
    	
	\maketitle
	
	\section{Introduction}
	\label{sec::introduction}
	The performance of deep learning models in remote sensing (RS) is heavily limited by the scarcity of high-quality labeled samples, which are costly and time-consuming to collect, often requiring expert knowledge. In contrast, Earth observation satellites continuously generate vast amounts of unlabeled imagery. Thus, a central challenge is to develop learning paradigms that effectively mitigate label scarcity by extracting greater utility from the abundant unlabeled data. Addressing this challenge is critical for enabling efficient and scalable applications in RS tasks such as scene classification\cite{cheng2017remote,wang2020looking,huang2023semi}, land-use classification\cite{karra2021global,wang2023review}, object detection\cite{li2017rotation,zhang2023efficient,zhang2024structured}, and semantic segmentation\cite{diakogiannis2020resunet,hua2021semantic,huang2024decouple}.
	
	To this end, \textbf{semi-supervised learning} (\textbf{SSL})\cite{oliver2018realistic,sohn2020fixmatch,zhang2021flexmatch,wang2023freematch} and \textbf{active learning}(\textbf{AL})\cite{settles2009active,yoo2019learning,bald,coreset,badge} offer complementary yet independently limited strategies. SSL enhances model performance by utilizing unlabeled data for model self-training, but its effectiveness depends significantly on the quality and representativeness of the initial labeled set, which is often randomly constructed and may lead to suboptimal learning. AL, on the other hand, mitigates annotation scarcity by iteratively selecting the most informative samples based on criteria such as diversity or representativeness for human labeling. However, by focusing only on a small subset of instances in each cycle, conventional AL fails to utilize the vast majority of unlabeled data, which remain unused in model training.
	
	The integration of these two strategies gives rise to \textbf{semi-supervised active learning} (\textbf{SSAL})\cite{huang2021semi,guo2021semi,gao2020consistency}, a unified paradigm in which SSL and AL complement and reinforce each other in a synergistic loop. In SSAL, the SSL component leverages both labeled data via supervised learning and unlabeled data through pseudo-labeling to gradually enhance the model’s representation quality and uncertainty estimation. Following the SSL phase, the AL module identifies and queries the most informative unlabeled samples for manual annotation, thereby expanding the labeled set in a targeted and cost-effective manner.
	Importantly, the two modules interact bidirectionally: a more discriminative SSL-trained model provides more discriminative representation and more reliable uncertainty cues to guide AL toward better query decisions, while the carefully selected samples from AL, in turn, supply higher-value labels that further strengthen SSL training in the next round. This mutual reinforcement enables SSAL to continuously improve both model performance and querying effectiveness. Despite its strong potential to reduce annotation costs and improve data efficiency, SSAL remains largely underexplored in the RS domain.
	
	To advance the SSAL paradigm for RS, this paper proposes a novel \textbf{Hierarchical Semi-Supervised Active Learning (HSSAL)} framework that integrates SSL with a hierarchical active learning (HAL) strategy in a closed iterative loop.
	Within each iteration, the SSL component focuses on fully exploiting the unlabeled data through a weak-to-strong self-training regime: low-uncertainty samples are assigned high-confidence pseudo-labels derived from weakly augmented inputs and are then used to supervise their strongly augmented counterparts. This pseudo-label–driven consistency learning effectively enlarges the training set without additional annotation cost.
	Guided by the enhanced representations and uncertainty estimates, the HAL component executes a hierarchical querying process that maximizes overall sample \textbf{informativeness} by jointly considering \textbf{\textit{scalability}}, \textbf{\textit{diversity}}, and \textbf{\textit{uncertainty}}. Specifically, HAL achieves scalability through mini-batch partitioning of the unlabeled pool, enhances diversity via spectral clustering that uncovers locally coherent structures, and captures uncertainty through gradient-based uncertainty scoring, ultimately identifying the most informative samples for annotation.
	
	The core idea of HSSAL is not simply to rely on uncertainty but to assign different learning strategies to different parts of the unlabeled data. In the SSL stage, samples with low uncertainty and therefore high confidence are selected and used in pseudo-label based  self-training, enabling the model to make effective use of a large amount of unlabeled imagery without additional annotation cost. In contrast, the HAL stage aims to identify the most valuable samples based on overall informativeness, which jointly considers scalability, diversity, and uncertainty, and selects these samples for manual annotation before the next SSL round. 
	Beyond the introduction of the novel HAL module, HSSAL is designed to be compatible with various SSL techniques, yielding significant performance improvements over standalone SSL or AL methods. 
	To validate its effectiveness, extensive experiments were conducted on three benchmark remote sensing scene classification datasets, including UCM, AID, and NWPU-RESISC45. The results show that HSSAL consistently surpasses both SSL-only and AL-only baselines. Remarkably, it achieves over \textbf{95\%} of the fully supervised accuracy using only \textbf{8\%}, \textbf{4\%}, and \textbf{2\%} of the labeled training data on the UCM, AID, and NWPU-RESISC45 datasets, respectively, demonstrating its superior label efficiency and robust capability to fully exploit unlabeled data across different uncertainty levels.

	The main contributions of this work are summarized as:
	\begin{itemize}
		\item \textbf{A unified uncertainty-aware learning framework.} 
		We propose a unified Hierarchical Semi-Supervised Active Learning framework that systematically integrates SSL and AL to fully exploit unlabeled data across different uncertainty levels, achieving a more efficient utilization of unlabeled RS samples.
		
		\item \textbf{A novel hierarchical active learning strategy.}  
		We propose a novel \textbf{HAL} strategy that hierarchically integrates scalability, diversity, and uncertainty into the criterion of sample informativeness, thereby ensuring that the queried samples are both representative and highly informative.
		
		\item \textbf{Comprehensive validation and superior label efficiency.} 
		Extensive experiments on three benchmark remote sensing scene classification datasets demonstrate that HSSAL consistently outperforms SSL- and AL-only baselines with a minimal label budget.
	\end{itemize}

	\section{Related Work}
	In this section, we made a brief review on SSL, AL, and SSAL, covering their developments in both the general domain and the field of remote sensing.
	
	\subsection{Semi-Supervised Learning}
	\subsubsection{General Semi-Supervised Learning}
	SSL has gained increasing attention in the deep learning era, as it effectively alleviates the annotation burden when labeled data are scarce. Early deep learning–based SSL methods leveraged adversarial learning, particularly GANs\cite{souly2017semi,hung2018adversarial,mittal2019semi}, to align feature distributions between labeled and unlabeled data via a discriminator that minimizes prediction entropy and domain discrepancy. While effective to some extent, these adversarial approaches suffer from poor interpretability and unstable optimization.  
	
	Subsequently, consistency-based methods emerged as a more stable alternative, enforcing prediction invariance under input or feature perturbations\cite{ouali2020semi,liu2022perturbed,sohn2020fixmatch,chen2021semi}. By encouraging consistent outputs for augmented views of the same sample, they enhance representation robustness and generalization. However, consistency-based methods often lack effective pseudo-label refinement, limiting iterative improvement through high-confidence predictions.
	
	Recently, self-training–based methods have become dominant due to their strong performance and simplicity. These approaches generate pseudo-labels for unlabeled data and jointly optimize supervised and self-training losses. Representative methods include FixMatch\cite{sohn2020fixmatch}, which filters pseudo-labels with a high-confidence threshold; FlexMatch\cite{zhang2021flexmatch}, which adjusts class-wise thresholds dynamically; FreeMatch\cite{wang2023freematch}, which uses adaptive confidence estimation to reduce overfitting; SoftMatch\cite{chen2023softmatch}, which applies soft weighting for uncertain samples; and CGMatch\cite{cheng2025cgmatch}, which partitions unlabeled samples into easy, ambiguous, and hard subsets for selective regularization. These methods collectively highlight the importance of confidence- and uncertainty-aware learning in SSL.
	
	\subsubsection{Semi-Supervised Learning in Remote Sensing}
	SSL has also been widely applied in RS to alleviate the annotation burden on massive unlabeled imagery. For RS scene classification, Miao et al.\cite{miao2022semi} proposed a semi-supervised representation consistency Siamese network combining Involution-GAN feature extraction with a Siamese consistency loss. Ding et al.\cite{ding2024uncertainty} introduced uncertainty-aware contrastive learning to handle reliable and unreliable samples differently, improving multimodal classification. Liu et al.\cite{liu2023multi} proposed a multi-level label-aware framework  in unlabeled data for better pseudo-labeling and feature discrimination. Huang et al.\cite{huang2023semi} developed a semi-supervised domain adaptation method for cross-domain RS scene classification, aligning source and target domains with both unsupervised and class-level supervised strategies. These advances improve RS-SSL by addressing pseudo-label noise, domain variability, and model uncertainty.
	
	Beyond scene classification, SSL has been explored in change detection\cite{bandara2022revisiting,zhang2023joint,shen2024learning}, road/building extraction\cite{huang2023adaptmatch,huang2024semi}, and RS semantic segmentation\cite{sun2020bas,zhang2022semi,chen2022semi,huang2024decouple,sun2025rsprotosemiseg,wang2025semi}, highlighting RS-specific challenges such as high intra-class diversity and inter-class similarity caused by heterogeneous ground and observation conditions.
	
	\subsection{Active and Semi-Supervised Active Learning}
	AL aims to enhance model performance under limited supervision by selectively querying the most informative samples for annotation, thereby maximizing the efficiency of labeling resources. Traditional AL methods generally rely on uncertainty-based or diversity-based sampling strategies. Uncertainty-based approaches\cite{lewis1995sequential,settles2009active,joshi2009multi} select samples on which the model is least confident, often measured via prediction entropy or margin sampling. Diversity-based methods\cite{coreset,badge,agarwal2020contextual} seek to ensure representativeness by covering diverse regions of the feature space, mitigating redundancy among queried samples. Recent deep AL frameworks further integrate these criteria with modern neural architectures: for instance, CoreSet\cite{coreset} formulates AL as a core-set selection problem via k-center clustering; BADGE\cite{badge} combines uncertainty and diversity by selecting gradient-embedding-based samples; and VAAL\cite{sinha2019variational} employs adversarial variational autoencoders to discriminate labeled and unlabeled distributions for sample selection. Uncertainty-aware  extension\cite{yoo2019learning} introduces deep uncertainty estimation and achieves improved performance with fewer labeled samples.
	
	In the RS domain, AL has been extensively adopted to alleviate the high annotation cost associated with large-scale satellite and aerial imagery. 
	Early studies\cite{tuia2009active,persello2014active,stumpf2013active,demir2010batch} primarily focused on uncertainty- or region-based querying for hyperspectral image (HSI) and land-cover classification, effectively identifying the most informative samples under limited labeling budgets. 
	To further enhance the reliability of sample selection, subsequent works incorporated spatial–spectral context\cite{patra2017spectral,tuia2011survey} and multi-view representations\cite{xu2021multiview} to better handle the heterogeneity of RS data. 
	With the advent of deep learning, AL has evolved toward feature-aware and structure-guided strategies. 
	For instance, Liu et al.\cite{lei2021active} proposed a deep AL framework for HSI classification that integrates spectral–spatial uncertainty with manifold regularization, achieving superior generalization under scarce supervision. 
	These advances collectively demonstrate that AL can substantially reduce labeling costs across diverse RS applications, including HSI analysis\cite{liu2018superpixel,cai2021phase}, scene classification\cite{stumpf2013active,demir2010batch}, and semantic segmentation\cite{lenczner2022dial,desai2022active}.

	To exploit the remaining unlabeled data more effectively, SSAL unifies AL’s selective querying with SSL’s ability to reduce the uncertainty of unlabeled samples, thereby achieving superior label efficiency. 
	Representative examples in the general domain include ConsistencyAL\cite{gao2020consistency}, which enforces prediction consistency between labeled and unlabeled data to guide query selection; SemiAL\cite{guo2021semi}, which alternates between pseudo-label–based self-training and uncertainty-driven annotation; and S$^3$AL\cite{huang2021semi}, which jointly optimizes sampling and model learning in a single semi-supervised objective. Compared with prior SSAL frameworks such as ConsistencyAL, SemiAL, and S$^3$AL that mainly rely on flat uncertainty- or consistency-based querying, HSSAL employs a hierarchical selection mechanism that jointly enforces scalability, diversity, and uncertainty, which is particularly effective for RS scene classification with highly non-convex and imbalanced class manifolds.

	\section{Hierarchical Semi-Supervised Active Learning}
	\label{sec::methodology}
	
	\subsection{Overview of HSSAL with Dataset Notations}
	\label{subsec::framework}
	
\begin{figure*}[h]
		\centering
		\includegraphics[width=1.0\textwidth]{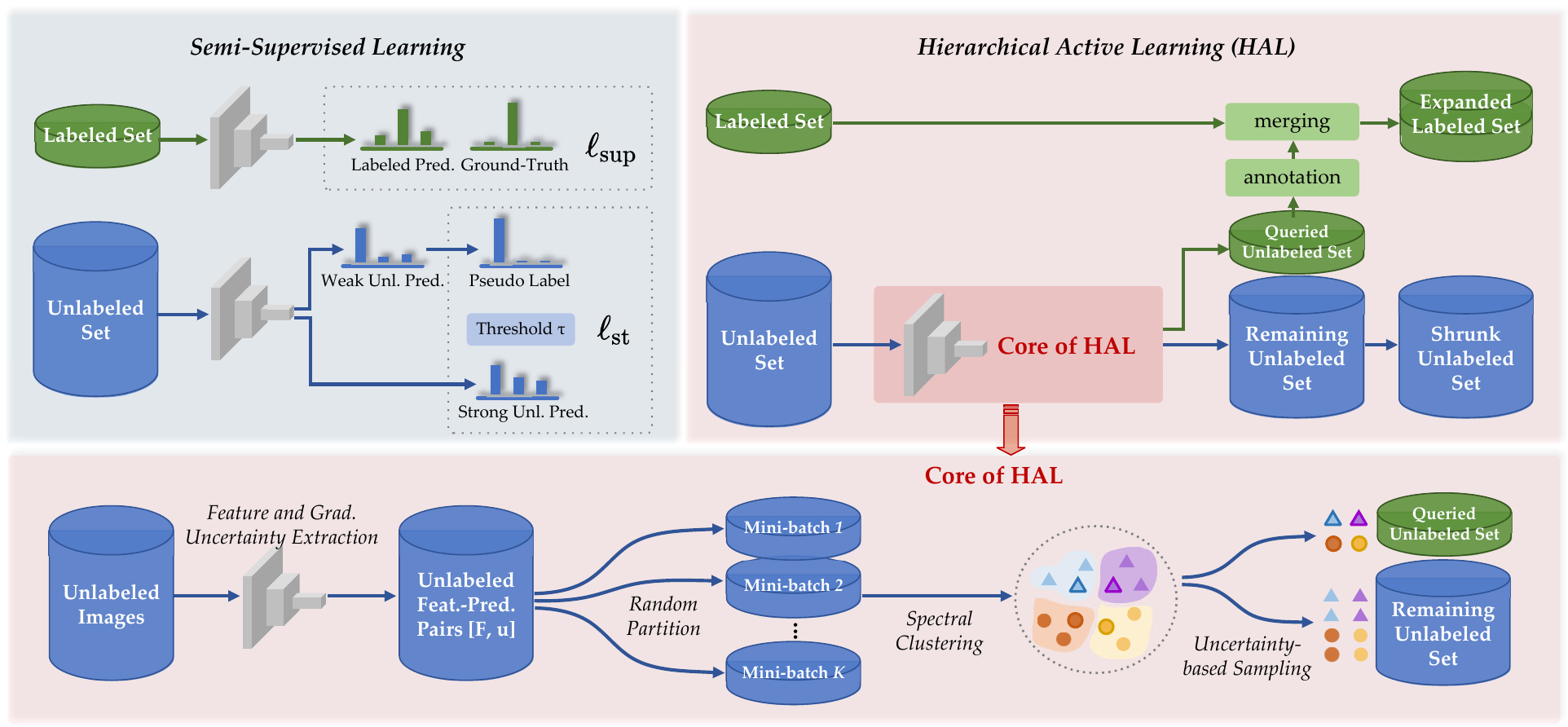}
		\caption{Overview of \textbf{HSSAL} for one round. Each round alternates \textbf{SSL} and \textbf{HAL} in a cooperative loop. 
			\textbf{SSL:} the model learns on labeled data via supervised loss and on unlabeled data via weak-to-strong self-training.
			\textbf{HAL:} systematically integrates \textit{\textbf{scalability}}, \textit{\textbf{diversity}}, and \textit{\textbf{uncertainty}} into a unified sample selection pipeline. It proceeds through four structured steps: (1) feature extraction and uncertainty estimation; (2) mini-batch partitioning for scalable processing; (3) spectral clustering with uncertainty-aware sampling; and (4) sample annotation and dataset update. 
			By hierarchically organizing these steps, HAL achieves scalable computation, diverse representation coverage, and uncertainty-driven querying. 
			Through iterative SSL–HAL interaction, HSSAL progressively refines both feature representations and labeled sets, exploiting low-uncertainty samples for learning and high-uncertainty ones for querying.}
		\label{fig:HSSAL}
	\end{figure*}

	The proposed \textbf{HSSAL} framework operates in an iterative SSL–HAL loop to achieve efficient model training and sample selection under minimal annotations.  
	An overview of HSSAL is shown in Fig.~\ref{fig:HSSAL}, and the detailed workflow for each round is summarized in Algorithm~\ref{alg::hssal}.
	
	Formally, HSSAL starts with a small labeled pool $\mathcal{L}^0 = \{(\mathbf{x}_i, y_i)\}_{i=1}^{M}$ and a large unlabeled pool $\mathcal{U}^0 = \{\mathbf{x}_j\}_{j=1}^{N}$, where $M \ll N$.  
	A model $f_{\theta}$ is iteratively trained over $R$ rounds, alternating between SSL-based model training and HAL-based sample selection.
	
	Across the entire training process, HSSAL proceeds for multiple iterative rounds, alternating between SSL-based model training and HAL-based sample querying. 
	In each round $r$, SSL trains the model on the current labeled and unlabeled sets, $\mathcal{L}^{r-1}$ and $\mathcal{U}^{r-1}$, by combining supervised learning on labeled data with weak-to-strong self-training on unlabeled data. 
	After convergence, HAL evaluates all unlabeled samples in $\mathcal{U}^{r-1}$ and selects an informative subset $\mathcal{Q}^r$ for annotation, jointly considering \textit{\textbf{scalability}}, \textit{\textbf{diversity}}, and \textit{\textbf{uncertainty}}. 
	To realize multi-round active learning, a predefined sequence of labeling ratios $\{\rho_1, \rho_2, \dots, \rho_R\}$ is introduced, where each $\rho_r$ denotes the cumulative proportion of labeled samples after round $r$. 
	Accordingly, the querying budget of round $r$ is defined as 
	$B_r = (\rho_r - \rho_{r-1}) |\mathcal{U}^{r-1}|$, 
	representing the number of newly annotated samples added in this iteration. 
	Within HAL, we denote $B = B_r$ and further distribute this budget across mini-batches for hierarchical sample selection. 
	The newly annotated subset $\mathcal{Q}^{r,l}$ is then incorporated into the labeled set $\mathcal{L}^r = \mathcal{L}^{r-1} \cup \mathcal{Q}^{r,l}$, 
	while the remaining unlabeled samples constitute $\mathcal{U}^r = \mathcal{U}^{r-1} \setminus \mathcal{Q}^r$. 
	This iterative design enables a controlled and balanced expansion of the labeled dataset throughout the learning process.
	
	This iterative process continues until the annotation budget is exhausted or model performance converges, forming a closed SSL–HAL loop that progressively identifies the most informative samples for annotation.  
	In the following subsections, we focus on the methodology within a single loop; for simplicity, the superscript \(r\) indicating the iteration index is omitted throughout.

	\begin{algorithm}[t]
		\caption{HSSAL for Each Round}
		\label{alg::hssal}
		\begin{algorithmic}[1]
			\REQUIRE labeled pool $\mathcal{L}$, unlabeled pool $\mathcal{U}$, total budget $B$, cluster multiplier $\lambda$, mini-batch size $S$
			\ENSURE Expanded labeled pooled $\mathcal{L}$
			\STATE Initialize model $\theta$, EMA model $\hat{\theta}$
			\textbf{\color{blue!70!black}\STATE {\texttt{// --- Phase A: Semi-supervised Learning --- //}}}
			\STATE Re-initialize model parameters;
			\FOR{epoch $= 1$ to $N_{\text{epochs}}$}
			\STATE \(\bullet\) Train $\theta$ on $\mathcal{L}$ and $\mathcal{U}$ using SSL strategy;
			\STATE \(\bullet\) Update EMA model $\hat{\theta}$ every iteration;
			\STATE \(\bullet\) Evaluate and save $\hat{\theta}$ based on validation set;
			\ENDFOR
			\textbf{\color{red!70!black}\STATE {\texttt{// --- Phase B: Hierarchical Active Learning --- //}}}
			\STATE {\textcolor{red!70!black}{\textit{Step I}}: Extract features $\mathbf{F}$ and uncertainties $\mathbf{u}$ for all samples in $\mathcal{U}$ using $\hat{\theta}$;}
			\STATE {\textcolor{red!70!black}{\textit{Step II}}: Randomly shuffle and divide $\mathcal{U}$ into batches $\{\mathcal{U}_b\}$, and allocate sample budget $B_b$ for each batch;}
			\STATE {\textcolor{red!70!black}{\textit{Step III}}: For each mini-batch $\mathcal{U}_b$, perform Spectral Clustering and select the most uncertain sample from each of the top-$B_b$ clusters ranked by total uncertainty;}
			\STATE {\textcolor{red!70!black}{\textit{Step IV}}: Annotate selected unlabeled samples $\mathcal{Q}$ to $\mathcal{Q}^l$, and update $\mathcal{U} \gets \mathcal{U} \setminus \mathcal{Q}$, $\mathcal{L} \gets \mathcal{L} \cup \mathcal{Q}^l$.}
			
			\RETURN Expanded labeled pool $\mathcal{L}$ and shrunken unlabeled pool $\mathcal{U}$ 
		\end{algorithmic}
	\end{algorithm}
	
	\subsection{Semi-Supervised Learning}
	\label{subsec::ssl_component}
	
	The SSL component in HSSAL is built upon a weak-to-strong self-training mechanism originating from FixMatch\cite{sohn2020fixmatch}, which enables the model to effectively leverage the large pool of unlabeled data by generating and refining pseudo-labels for self-supervised training. This mechanism represents one of the most classical SSL paradigms, allowing the model to jointly learn from a small set of labeled samples and a large set of pseudo-labeled samples produced from its own predictions under weak augmentations.
	
	\paragraph{\textbf{Supervised Learning}.}
	Given the labeled dataset $\mathcal{L} = \{(\mathbf{x}_i, y_i)\}_{i=1}^{M}$, the supervised loss $\ell_{\text{sup}}$ is computed using the standard cross-entropy loss:
	\begin{equation}
		\label{eq::sup_loss}
		\ell_{\text{sup}} = 
		-\log p_{\theta}\!\left(y = y_i \mid \mathbf{x}_i\right),
	\end{equation}

	\paragraph{\textbf{Weak-to-strong-based Self-Training}.}
	For each unlabeled image $\mathbf{x}_j \in \mathcal{U}$, a weakly augmented version $\mathbf{x}_j^{w}$ and a strongly augmented version $\mathbf{x}_j^{s}$ are generated using two different augmentation functions $\mathcal{A}_{w}$ and $\mathcal{A}_{s}$ respectively, with $\mathcal{A}_{s}$ being more aggressive. The model first predicts a class distribution on the weakly augmented input:
	\[
	p_j^{w} = p_{\hat{\theta}}(y|\mathbf{x}_j^{w}),
	\]
	where $\hat{\theta}$ denotes the Exponential Moving Average (EMA) model that provides more stable pseudo-labels.  
	
	The pseudo-label $\hat{y}_j$ is obtained as the class with the highest probability:
	\[
	\hat{y}_j = \arg\max_{c} p_j^{w}(c),
	\]
	and its confidence $\max_c p_j^{w}(c)$ is used to determine whether the sample should contribute to the unsupervised loss.  
	To ensure robustness across different SSL variants (\textit{e.g.}, FixMatch, FlexMatch, FreeMatch, SoftMatch), 
	we introduce a unified confidence mask $m_j$ that filters/weights uncertain pseudo-labels with threshold $\tau$ as:
	\begin{equation}
		\label{eq::mask}
		m_j = \mathbb{1}\!\left(\max_c p_j^{w}(c) \ge \tau\right).
	\end{equation}
	The self-training objective for the unlabeled samples, \textit{i.e.}, the unsupervised loss component, is defined as follows:
	\begin{equation}
		\label{eq::st_loss}
		\ell_{\text{st}} = 
		-m_j \, \log p^s_{\theta}\!\left(y = \hat{y}_j \mid \mathbf{x}_j^{s}\right),
	\end{equation}
	where $p_j^s$ is the prediction of strongly-augmented unlabeled sample as $p_j^{s} = p_{\hat{\theta}}(y|\mathbf{x}_j^{s})$, and $m_j$ is illustrated as a binary confidence mask. When instantiating HSSAL with specific SSL algorithms, $m_j$ can be either a hard indicator (\textit{e.g.}, FixMatch, FlexMatch, FreeMatch) or a soft confidence weight (\textit{e.g.}, SoftMatch), following their original designs.
	
	In other words, the model is encouraged to produce consistent predictions for the strongly augmented sample $\mathbf{x}_j^{s}$, guided by the pseudo-label derived from its weakly augmented counterpart $\mathbf{x}_j^{w}$. 
	Low-confidence samples ($<\tau$) are excluded to suppress noisy pseudo-labels. 
	When instantiating HSSAL with specific SSL algorithms (\textit{e.g.}, FixMatch, FlexMatch, FreeMatch, or SoftMatch), we adopt their original thresholding or weighting strategies as particular realizations of the generic mask $m_j$ in Eq.~(\ref{eq::mask}), such as a fixed $\tau=0.95$ in FixMatch, class-adaptive $\tau_c$ in FlexMatch, or dynamically adjusted thresholds in FreeMatch.

	\paragraph{\textbf{Overall Objective}.}
	The total loss for each round combines the supervised and pseudo-labeled components:
	\begin{equation}
		\label{eq::ssl_total_loss}
		\ell_{\text{total}} = \ell_{\text{sup}} + \lambda_{st}\ell_{\text{st}},
	\end{equation}
	where $\lambda_{st}$ is the self-training weight and its value depends on certain SSL methods.
	
	The EMA model $\hat{\theta}$ is updated after each training iteration to stabilize training and prevents the propagation of noisy pseudo-labels, using a momentum coefficient $\alpha$:
	\[
	\hat{\theta} \leftarrow \alpha \hat{\theta} + (1 - \alpha)\theta.
	\]
	
	The SSL improves uncertainty estimation by leveraging low-uncertainty unlabeled data for self-training, allowing the model to focus on the real high-uncertainty unlabeled examples and providing a solid foundation for subsequent active querying.

	\subsection{Hierarchical Active Learning}
	This section introduces the \textbf{HAL} framework, which systematically incorporates \textit{\textbf{scalability}}, \textit{\textbf{diversity}}, and \textit{\textbf{uncertainty}} into a unified sample selection pipeline.
	As shown in \textit{Phase~B} of Algorithm~\ref{alg::hssal}, HAL proceeds through four structured steps: (1) feature extraction and uncertainty estimation, (2) mini-batch partitioning for scalable processing, (3) spectral clustering with uncertainty-aware sampling, and (4) sample annotation and dataset update.
	By hierarchically organizing these steps, HAL achieves scalable computation, diverse representation coverage, and uncertainty-driven querying, ensuring both efficiency and informativeness on large-scale RS datasets.
	
	\subsubsection{Step~I: Feature Extraction and Uncertainty Estimation}
	We decompose the EMA model into an encoder–classifier architecture with parameters $\hat{\theta} = (\hat{\theta}_{\text{enc}}, W)$, where $\hat{\theta}_{\text{enc}}$ denotes the DINOv2-Small encoder and $W$ is the final linear classifier.
	HAL leverages the encoder features for clustering and the classifier gradients for uncertainty estimation, following the principle of BADGE\cite{badge} but without performing gradient-space clustering. 
	Building upon the well-trained EMA model obtained from the preceding SSL phase, HAL characterizes the entire unlabeled set $\mathcal{U}=\{\mathbf{x}_i\}_{i=1}^{N}$ to derive two complementary representations: 
	a global feature matrix $\mathbf{F}$ that captures the semantic distribution, and an uncertainty vector $\mathbf{u}$ that quantifies sample informativeness. 
	These two components jointly form the basis for the subsequent mini-batch partitioning and uncertainty-aware querying process.
	
	\paragraph{(A) \textbf{Feature Extraction}.}
	Each image $x_i$ in the unlabeled set is processed by the EMA version of the DINOv2-Small encoder, parameterized by $\hat{\theta}_{\text{enc}}$ (well-trained in the SSL phase), to extract a compact $D$-dimensional ($D=384$) global feature representation. The final-layer \texttt{[CLS]} token yields a global image embedding:
	\begin{equation}
		\mathbf{f}_i = f(x_i; \hat{\theta}_{\text{enc}}) \in \mathbb{R}^{D},
	\end{equation}
	which captures rich global semantic information of the RS scene image. All feature vectors are stacked to form the feature matrix:
	\begin{equation}
		\mathbf{F} = [\mathbf{f}_1, \mathbf{f}_2, \dots, \mathbf{f}_N]^\top \in \mathbb{R}^{N \times D}.
	\end{equation}
	where \(N\) denotes the total number of samples.

	\paragraph{(B) \textbf{Uncertainty Estimation}.}
	Unlike conventional entropy-based or margin-based methods, we adopt the gradient norm of the classifier as the uncertainty metric, following\cite{badge}. For each sample $x_i$, the uncertainty score $u_i$ is computed as:
	\begin{equation}
		u_i = \left\| \nabla_{W} \mathcal{L}_{CE}(f(x_i; \hat{\theta}), \hat{y}_i) \right\|_2,
	\end{equation}
	where $\hat{y}_i = \arg\max f(x_i; \hat{\theta})$ denotes the pseudo-label generated by the EMA model, and $W$ represents the weight matrix of the last fully-connected layer.
	
	The gradient computation proceeds as follows: first, we compute the CE loss between the model prediction and the pseudo-label; then, we perform backward propagation to calculate the gradient with respect to the classifier weights; finally, we compute the $\ell_2$-norm of the flattened gradient vector.
	A larger gradient norm $u_i$ indicates that the model's loss landscape is steep around $x_i$, suggesting that $x_i$ lies near a decision boundary with high uncertain.
	
	All uncertainty scores are aggregated into an vector:
	\begin{equation}
		\mathbf{u} = [u_1, u_2, \dots, u_N]^\top \in \mathbb{R}^{N},
	\end{equation}
	which together with the feature matrix $\mathbf{F}$, forms the pair $\{\mathbf{F}, \mathbf{u}\}$ representing both the feature embeddings and uncertainty of $\mathcal{U}$.
	
	\subsubsection{Step~II: Mini-batch Partition for Scalable Selection}
	Spectral clustering on the entire unlabeled pool $\mathcal{U}$ is computationally demanding due to its super-quadratic complexity in the number of samples. To address this scalability challenge, HAL employs a mini-batch partitioning strategy that decomposes the global selection problem into tractable subproblems.

	Specifically, HAL partitions $\mathcal{U}$ into $K$ batches $\{\mathcal{U}_b\}_{b=1}^{K}$ by specifying a fixed mini-batch size $S$, where $K = \lceil N/S \rceil$ represents the total number of batches. The last batch $\mathcal{U}_K$ contains the remaining $N \bmod S$ samples when $N$ is not divisible by $S$. This partitioning induces corresponding splits:
	\begin{equation}
		\mathbf{F} = [\mathbf{F}_1, \mathbf{F}_2, \dots, \mathbf{F}_K], \quad \mathbf{u} = [\mathbf{u}_1, \mathbf{u}_2, \dots, \mathbf{u}_K],
	\end{equation}
	where $\mathbf{F}_b \in \mathbb{R}^{|\mathcal{U}_b| \times D}$ and $\mathbf{u}_b \in \mathbb{R}^{|\mathcal{U}_b|}$ contain the features and uncertainty scores for batch $\mathcal{U}_b$.
	
	The annotation budget is distributed proportionally across batches:
	\begin{equation}
		B_b = 
		\begin{cases} 
			\left\lfloor B \cdot \frac{|\mathcal{U}_b|}{|\mathcal{U}|} \right\rfloor, & b = 1, \dots, K-1, \\
			B - \sum_{j=1}^{K-1} B_j, & b = K, 
		\end{cases}
	\end{equation}
	ensuring $\sum_{b=1}^{K} B_b = B$.
	
	\subsubsection{Step~III: Spectral Clustering and Uncertainty-aware Sampling}
	After obtaining feature–uncertainty pairs for all batches, HAL performs spectral clustering and uncertainty-aware sampling within each mini-batch to query most informative samples.
	
	\paragraph{(A) \textbf{Spectral Clustering}.}
	For each mini-batch $\mathcal{U}_b$ containing $N_b$ ($N_b>1$) unlabeled samples, the number of spectral clusters is adaptively determined as:
	\begin{equation}
		K_b = \min(\lambda B_b,\, N_b - 1),
		\label{eq:num_clusters}
	\end{equation}
	where $B_b$ is the allocated annotation budget for mini-batch $\mathcal{U}_b$, and $\lambda>1$ is a cluster multiplier that ensures the number of clusters exceeds the selection quota, thereby promoting fine-grained diversity during sampling.  
	
	The clustering consists of four sequential sub-steps.
	
	\textit{\textbf{(1) Graph Construction.}}  
	Given the feature matrix $\mathbf{F}_b = [\mathbf{f}_1, \mathbf{f}_2, \dots, \mathbf{f}_{N_b}]^\top \in \mathbb{R}^{N_b \times D}$,    
	a $k$-nearest neighbor (kNN) graph is built to model pairwise similarity:
	\begin{equation}
		\mathbf{A} = \text{KNN}(\mathbf{F}_b, k), \quad 
		\mathbf{A}_{\text{sym}} = \max(\mathbf{A}, \mathbf{A}^\top),
	\end{equation}
	where $\mathbf{A}$ is the directed adjacency matrix and $\mathbf{A}_{\text{sym}}$ is its symmetrized version,  
	ensuring an undirected graph suitable for Laplacian computation.  
	A fixed neighbor size of $k=40$ is used in this study.

	\textit{\textbf{(2) Laplacian Normalization.}}
	Let $\mathbf{D}$ be the degree matrix with diagonal entries $D_{ii} = \sum_j A_{\text{sym},ij}$.  
	The normalized graph Laplacian is then computed as:
	\begin{equation}
		\mathbf{L} = \mathbf{I} - \mathbf{D}^{-1/2}\mathbf{A}_{\text{sym}}\mathbf{D}^{-1/2},
	\end{equation}
	where $\mathbf{I}$ is the identity matrix of size $N_b \times N_b$.
	
	\textit{\textbf{(3) Spectral Embedding.}}
	The eigenvalue problem
	\begin{equation}
		\mathbf{L}\mathbf{v}_i = \lambda_i\mathbf{v}_i, \quad i = 1, \dots, K_b,
	\end{equation}
	is solved for the $K_b$ smallest eigenvalues $(\lambda_1 \le \lambda_2 \le \cdots \le \lambda_{K_b})$ and their corresponding eigenvectors $\mathbf{v}_i \in \mathbb{R}^{N_b}$.  
	The spectral embedding is defined as:
	\begin{equation}
		\mathbf{E} = [\mathbf{v}_1, \mathbf{v}_2, \dots, \mathbf{v}_{K_b}] \in \mathbb{R}^{N_b \times K_b},
	\end{equation}
	which provides a compact manifold representation of the data in a low-dimensional subspace. Before clustering, the rows of $\mathbf{E}$ are $\ell_2$-normalized as $\tilde{\mathbf{E}}$ to unit length.
	
	\textit{\textbf{(4) Clustering.}}
	Finally, $k$-means is performed on the rows of $\tilde{\mathbf{E}}$ to obtain $K_b$ disjoint clusters:
	\begin{equation}
		\{\mathcal{C}_1, \dots, \mathcal{C}_{K_b}\}, \quad \bigcup_{i=1}^{K_b}\mathcal{C}_i = \mathcal{U}_b, \quad \mathcal{C}_i \cap \mathcal{C}_j = \emptyset.
	\end{equation}
	Each cluster $\mathcal{C}_i$ corresponds to a group of samples sharing similar spectral representations in the feature space.  
	Unlike conventional distance-based clustering methods (\textit{e.g.}, $k$-means in the original feature space), spectral clustering operates on the eigenstructure of the graph Laplacian, enabling it to capture nonlinear manifolds and complex similarity relationships among samples.  
	This property is particularly advantageous for RS imagery, where diverse and spatially mixed land-cover types create highly non-convex class boundaries across heterogeneous feature manifolds.
	
	\paragraph{(B) \textbf{Uncertainty-aware Sampling.}}
	\label{sec::uncertainty_sampling}
	After spectral clustering, HAL performs uncertainty-guided sampling to jointly ensure informativeness and diversity.
	For each mini-batch $\mathcal{U}_b$, the uncertainty of each cluster $\mathcal{C}_j$ is aggregated as
	\begin{equation}
		U_j = \sum_{i \in \mathcal{C}_j} u_i,
	\end{equation}
	which measures the overall informativeness of its member samples.
	Clusters are ranked in descending order of $U_j$, and the top-$B_b$ clusters are retained for sample selection.
	From each selected cluster $\mathcal{C}_j$, the most uncertain sample is chosen as its representative:
	\begin{equation}
		i_j^\star = \operatorname*{arg\,max}_{i \in \mathcal{C}_j} u_i .
	\end{equation}
	
	This hybrid strategy ensures that selected samples are both highly informative and spatially diverse, effectively reducing redundancy while maintaining balanced coverage of the feature space.
	
	\subsubsection{Step~IV: Sample Annotation and Dataset Update}
	Once the informative and diverse samples are selected through the hierarchical querying process, HAL proceeds to the annotation and dataset update stage to complete the AL loop and prepare for the next semi-supervised training round.
	
	After the uncertainty-aware sampling within each batch, the selected indices are aggregated as the batch-level query set:
	\begin{equation}
		\mathcal{Q}_b = \{\, i_j^\star \mid j = 1, \dots, B_b \,\}.
	\end{equation}
	All batch-level queries are then merged to form the global query set:
	\begin{equation}
		\mathcal{Q} = \bigcup_{b=1}^{K} \mathcal{Q}_b .
	\end{equation}
	
	The samples in $\mathcal{Q}$ are manually annotated to obtain the corresponding labeled set $\mathcal{Q}^l$.
	After annotation, the labeled and unlabeled pools are updated as
	\begin{equation}
		\mathcal{L} \leftarrow \mathcal{L} \cup \mathcal{Q}^l, \qquad
		\mathcal{U} \leftarrow \mathcal{U} \setminus \mathcal{Q}.
	\end{equation}

	The expanded labeled dataset $\mathcal{L}$ serves as the input for the next SSL phase, thereby forming a closed AL loop.
	By alternating between SSL-based model training and HAL-based sample querying, the framework progressively improves model generalization while minimizing manual annotation cost.
	
	\subsubsection{Advantages of HAL}
	Overall, HAL boosts the informativeness of queried samples from three perspectives:
	\begin{itemize}
		\item \textbf{\textit{Scalability}:} mini-batch partitioning enables efficient large-scale processing.
		\item \textbf{\textit{Diversity}:} spectral clustering ensures selection across distinct feature manifolds.
		\item \textbf{\textit{Uncertainty}:} uncertainty-aware sampling focuses on selecting samples with high uncertainty.
	\end{itemize}
	By integrating global uncertainty analysis with local structure-aware selection, HAL achieves robust and efficient sample querying for RS scene classification.

	\section{Experiments}
	This section first introduces the used datasets, evaluation metrics, and experimental settings. It then presents and analyzes the results of the ablation studies, hyperparameter sensitivity experiments, clustering strategy comparisons, and comparison experiments of the proposed HSSAL framework and various SOTA SSL and AL methods.
	
	\subsection{Datasets, metrics, and experimental settings}
	\label{subsec::experimental_settings}
	
	\textbf{Datasets.} To comprehensively evaluate the effectiveness of the proposed HSSAL framework for RS scene classification, extensive experiments were conducted on three widely used benchmarks:
	(1) \textbf{UCM}\cite{UCM} contains 2,100 aerial images of size $256\times256$ pixels across 21 land-cover categories such as residential, forest, and airport scenes, with a spatial resolution of approximately 0.3 m;
	(2) \textbf{AID}\cite{AID} comprises 10,000 images of size $600\times600$ pixels collected from Google Earth, spanning 30 scene types with high intra-class variation and inter-class similarity, at a spatial resolution ranging from 0.5 m to 8 m; and
	(3) \textbf{NWPU-RESISC45}\cite{RESISC45} consists of 31,500 images of size $256\times256$ pixels over 45 scene classes, representing one of the largest and most diverse RS scene classification benchmarks to date, with spatial resolutions varying between 0.2 m and 30 m.
	All datasets provide RGB imagery under diverse spatial resolutions and geographic distributions, allowing for a comprehensive evaluation of the model’s generalization capability.
	
	\textbf{Metrics.} For quantitative evaluation, two widely adopted metrics were employed: Overall Accuracy (\textbf{OA}) and Average Accuracy (\textbf{\textbf{AA}}). OA measures the proportion of correctly classified samples across all classes, while AA computes the mean classification accuracy over individual classes to account for class imbalance.

	\textbf{Experimental Settings.}  
	All experiments, including the proposed HSSAL framework and comparison experiments, adopt the DINOv2-Small encoder\cite{oquab2023dinov2} combined with a lightweight classifier as the default classification model.
	The classifier consists of a Layer Normalization, a fully connected layer that reduces the embedding dimension to 128, followed by a GELU activation, a Dropout layer (0.1), and a final Linear layer mapping to the number of classes.
	The optimizer is AdamW, with the encoder learning rate initialized at $5\times10^{-6}$ and decayed following a polynomial schedule $(1 - iter/N_{iter})^{0.9}$.
	The classifier learning rate is set to 20$\times$ that of the encoder. 
	Labeling budgets of [1\%, 2\%, 4\%, 6\%, 8\%, 10\%] were applied in the iterative active learning cycle across all three datasets. In all cases, the 1\% budget provided the initial labeled set, which was shared among all methods and configurations.
	
	To meet the input requirements of the DINOv2-Small encoder, where the patch embedding resolution must be a multiple of 14, all images are resized to $252\times252$ pixels for training and inference.
	A batch size of 32 is used for all three datasets; the UCM dataset is trained for 20 epochs, while AID and NWPU-RESISC45 are each trained for 10 epochs.
	For each dataset, we fix a single 7:1:2 train/val/test split and, within the training set, a single random 1\% labeled subset as the initial pool, which is shared by all methods across all runs for fair comparison.
	During the SSL phase, an EMA model with a decay rate of 0.99 is maintained from the trainable version, and the EMA model achieving the best validation performance is retained for final evaluation on the test set.
	
	To enhance model generalization, weak augmentations are applied to both labeled and weakly augmented unlabeled images, including random vertical and horizontal flipping, random rescaling and cropping within the range [0.8, 1.0].  
	For the unlabeled branch, three additional strong augmentation strategies, including color jittering, random grayscaling, and Gaussian blurring, are further applied on top of the weak augmentations to obtain the strongly augmented versions.  
	
	\subsection{Hyperparameter Experiments of HAL}
	\label{subsec:hyperparam}
	
	This subsection investigates the key hyperparameters that influence the \textbf{\textit{scalability}}, \textbf{\textit{diversity}}, and \textbf{\textit{uncertainty}} of the proposed \textbf{HAL} module, while keeping all other training configurations identical to those described in Section~\ref{subsec::experimental_settings}. These include the AdamW optimizer, batch size, number of epochs, EMA decay rate of 0.99, and the DINOv2-Small encoder with the same classifier within the iterative loops under corresponding labeled budgets. All results are reported with OA and AA.
	
	\subsubsection{Effect of Mini-Batch Size (Scalability)}
	\label{subsubsec:hp_minibatch}
\begin{figure}[h]
		\centering
		\includegraphics[width=0.43\textwidth]{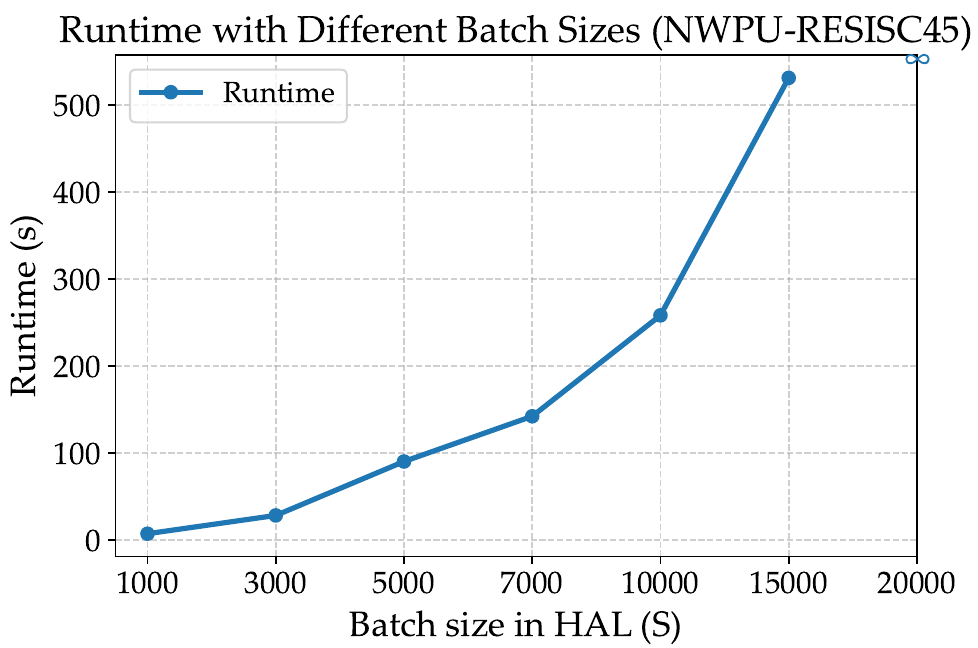}
		\caption{Runtime of HAL under different mini-batch sizes on 8\% labeled NWPU-RESISC45. Larger batches significantly increase runtime due to the super-quadratic complexity of spectral clustering.}
		\label{fig:running_time_batch_curve}
	\end{figure}

	To evaluate the scalability of the HAL for sample selection, we vary the mini-batch size $S$ used in the HAL stage on the large-scale NWPU-RESISC45, while keeping the rest of HAL intact. Specifically, we test $ S \in \{1000,\, 3000,\, 5000,\, $  $ 7000,\, 10000,$ $15000,\, 20000\}$. The cluster multiplier $\lambda$ in Eq. (\ref{eq:num_clusters}) is fixed as 3.0. We report OA/AA and runtime to characterize the selection quality and computational cost. This experiment highlights HAL's scalability achieved via mini-batch partitioning. Results are summarized in The results are summarized in Table~\ref{tab:experiments_batch_size}, and the total runtime of spectral clustering for different unlabeled partition under the 8\% labeled ratio is illustrated in Fig.~\ref{fig:running_time_batch_curve}.
	
	\begin{table*}[h]
		\center
		\caption{Evaluation of HAL's scalability on NWPU-RESISC45 under different mini-batch sizes, reporting overall accuracy (OA), average accuracy (AA), and the runtime (seconds) required to perform spectral clustering on the entire unlabeled pool during each round. The best OA and AA at each labeling ratio (except 1\%) are highlighted in \textbf{bold}. Here, ``--'' indicates that no results are reported because the runtime was excessively long, causing the \texttt{sklearn} spectral clustering algorithm to terminate automatically. }
		\label{tab:experiments_batch_size}	
		\setlength{\extrarowheight}{0mm}
		\setlength\tabcolsep{3pt}
		\resizebox{1.00\textwidth}{!}{
			\begin{tabular}{c | c c c | c c c | c c c | c c c | c c c | c c }
				\hline			
				\hline		
				\multirow{2}*{Mini-Batch Size}
				& \multicolumn{3}{c|}{1\% (26730 unlabeled)}  
				& \multicolumn{3}{c|}{2\% (26460 unlabeled)}  
				& \multicolumn{3}{c|}{4\% (25920 unlabeled)}  
				& \multicolumn{3}{c|}{6\% (25380 unlabeled)}  
				& \multicolumn{3}{c|}{8\% (24840 unlabeled)}  
				& \multicolumn{2}{c}{10 \% (24300 unlabeled)} \\ 
				\cline{2-18}
				
				& OA & AA & Time & OA & AA & Time  
				& OA & AA & Time & OA & AA & Time 
				& OA & AA & Time & OA & AA \\
				\hline
				
				1000 
				& 73.6  & 73.6  & 3  
				& 79.4  & 86.2  & 7 
				& 90.1  & 90.1  & 7 
				& 92.2  & 92.2  & 7 
				& 93.9  & 94.0  & 7 
				& 94.3  & 94.3  \\ 
				
				3000 
				& 71.6  & 71.6  & 11  
				& 79.8  & 80.0  & 28  
				& 89.3  & 89.3  & 29  
				& 92.6  & 92.7  & 27  
				& 94.0  & 94.0  & 28  
				& 95.1  & 95.1  \\ 
				
				5000 
				& 71.3  & 71.5  & 23  
				& 83.0  & 83.1  & 78 
				& 89.8  & 89.8  & 82  
				& 93.1  & 93.2  & 90  
				& 93.6  & 93.6  & 90  
				& 94.7  & 94.7  \\ 
				
				7000 
				& 73.2  & 73.2  & 49  
				& 83.3  & 83.5  & 142  
				& 90.8  & 90.8  & 140  
				& 92.6  & 92.6  & 142  
				& \textbf{94.5}  & \textbf{94.6}  & 142  
				& \textbf{94.8}  & \textbf{94.8}  \\ 
				
				10000 
				& 72.4  & 72.5  & 96  
				& 84.5  & 84.4  & 262  
				& \textbf{91.0}  & \textbf{91.1}  & 256  
				& \textbf{93.4}  & \textbf{93.4}  & 258  
				& 93.7  & 93.8  & 258 
				& 94.7  & 94.7  \\ 
				
				15000 
				& 71.1  & 71.1  & 173  
				& \textbf{85.9}  & \textbf{86.1}  & 532  
				& 90.6  & 90.6  & 524  
				& 92.9  & 92.9  & 520  
				& 94.0  & 94.1  & 531  
				& 94.3  & 94.3  \\ 
				
				20000 
				& --  & --  & --  
				& --  & --  & --  
				& --  & --  & --  
				& --  & --  & --  
				& --  & --  & --  
				& --  & --   \\ 
				
				\hline \hline
			\end{tabular}
		}
	\end{table*} 	
	
	As shown in Table~\ref{tab:experiments_batch_size}, overall, increasing the mini-batch size from 1000 to 5000 improves both OA and AA across all labeling ratios while maintaining moderate runtime. The mini-batch strategy, which randomly partitions the unlabeled pool into multiple subsets and performs spectral clustering within each subset, effectively preserves selection quality and diversity while greatly reducing computational complexity. This design allows HAL to maintain both global representativeness and local diversity among queried samples, making spectral clustering practical even for large-scale unlabeled datasets.
	
	When the mini-batch size exceeds 7000, performance begins to fluctuate slightly, and the runtime increases sharply due to the super-quadratic complexity of spectral clustering. Considering the trade-off between accuracy and efficiency, $S=10000$ achieves the best overall balance, maintaining competitive OA/AA while keeping runtime within a feasible range, and it is adopted as the default configuration in subsequent experiments. 
	These results validate the \textit{\textbf{scalability}} of HAL’s batch-wise spectral clustering design, demonstrating its ability to retain robust selection performance while remaining computationally efficient on large-scale data.

	\subsubsection{Effect of Cluster Multiplier $\lambda$ (Diversity)}
	\label{subsubsec:hp_lambda}
\begin{figure*}[!t]
		\centering
		\begin{minipage}[b]{0.245\textwidth}
			\centering
			\includegraphics[width=\textwidth]{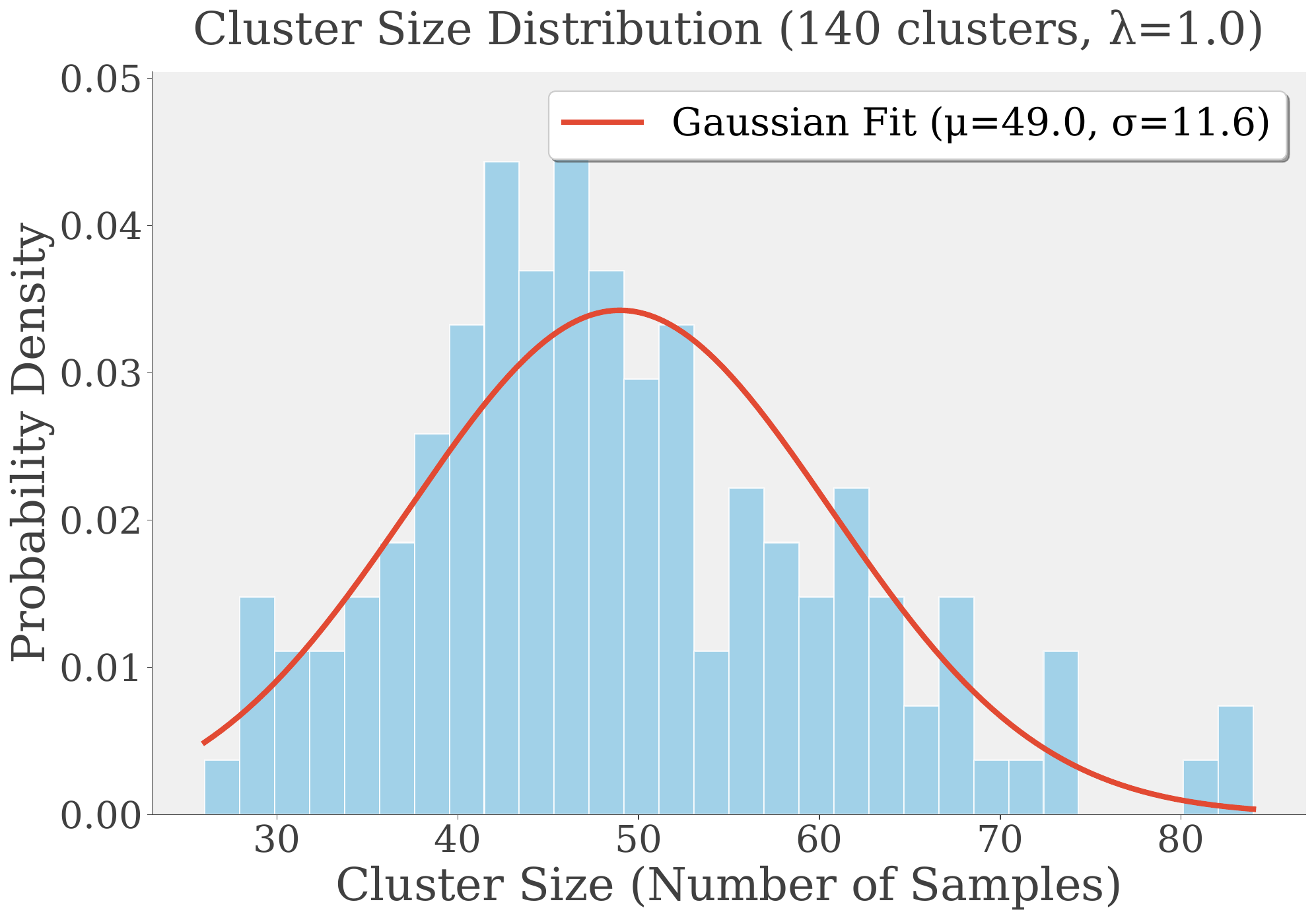}
			\label{fig:cluster_size_lambda1}
		\end{minipage}
		\begin{minipage}[b]{0.245\textwidth}
			\centering
			\includegraphics[width=\textwidth]{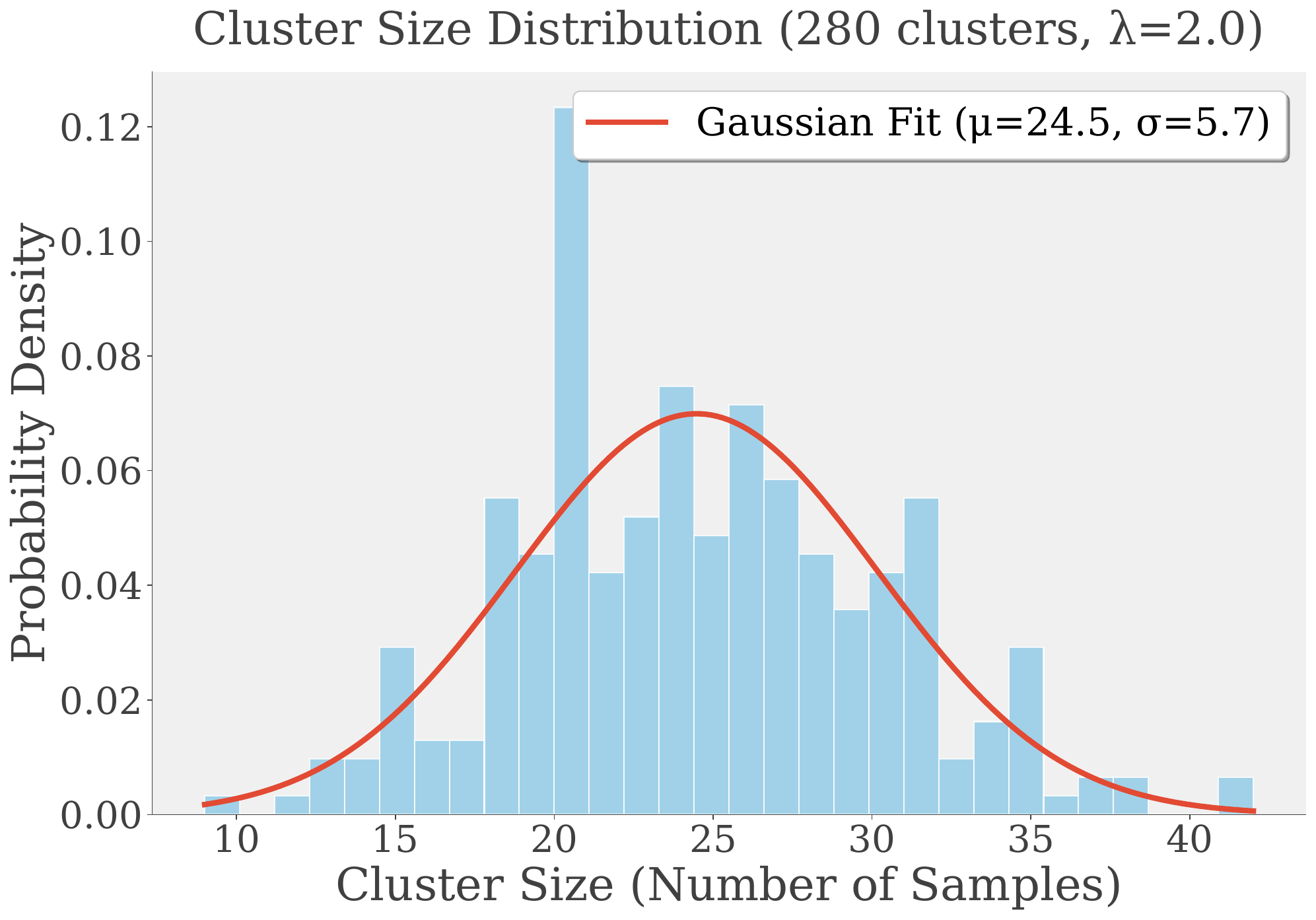}
			\label{fig:cluster_size_lambda2}
		\end{minipage}
		\begin{minipage}[b]{0.245\textwidth}
			\centering
			\includegraphics[width=\textwidth]{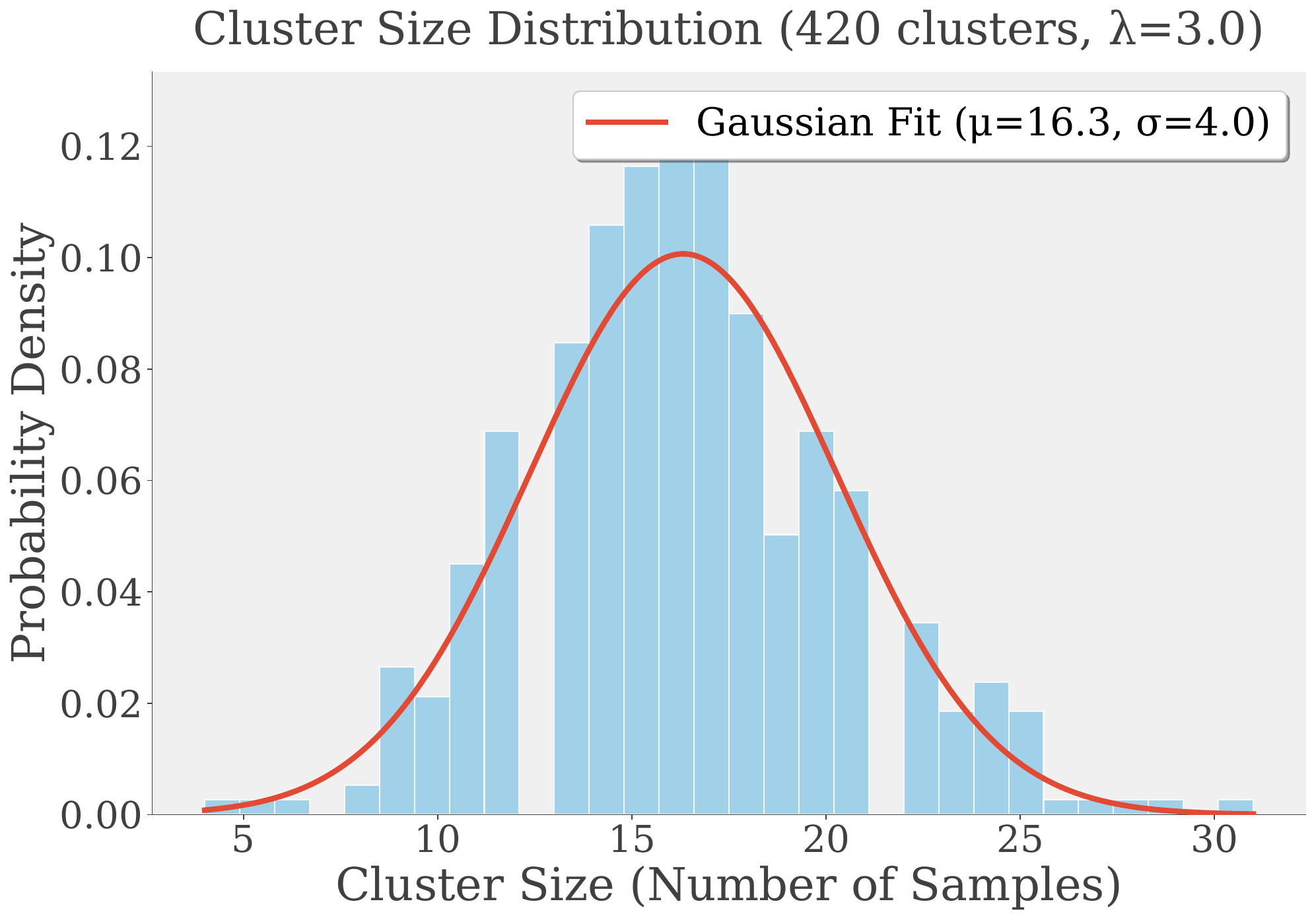}
			\label{fig:cluster_size_lambda3}
		\end{minipage}
		\begin{minipage}[b]{0.245\textwidth}
			\centering
			\includegraphics[width=\textwidth]{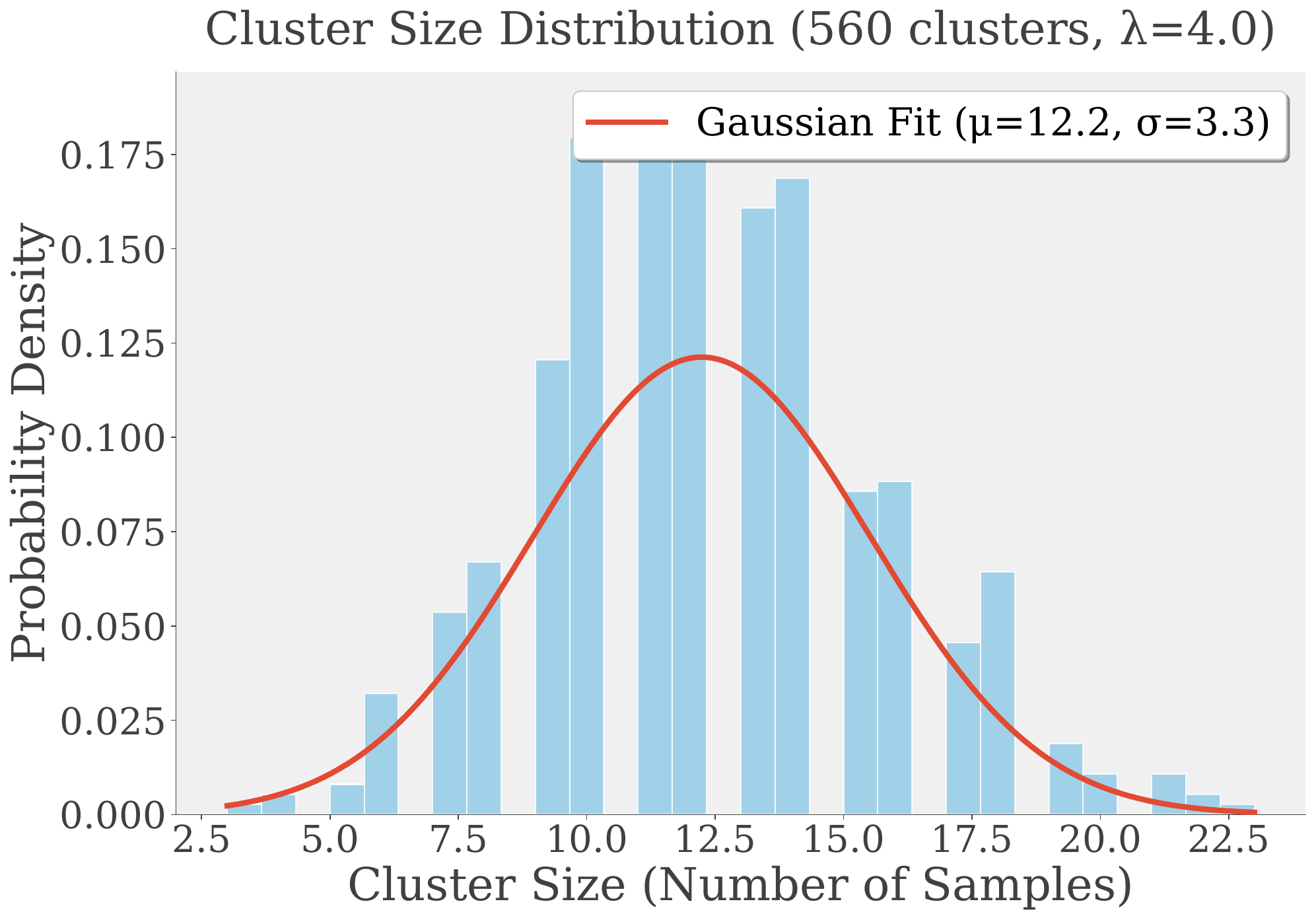}
			\label{fig:cluster_size_lambda4}
		\end{minipage}
		\vspace{-0.7cm}
		\caption{Distribution of spectral cluster sizes under different values of $\lambda$, which controls the number of clusters.}		
		\label{fig:cluster_size_distribution}
	\end{figure*}

	To investigate the impact of clustering granularity in the spectral clustering stage of HAL, we examine the cluster multiplier $\lambda$, which determines the number of spectral clusters in each mini-batch as $K_b = \lambda B_b$ in Eq. (\ref{eq:num_clusters}), where $B_b$ denotes the allocated query budget.
	By varying $\lambda$, we control the fineness of spectral clustering, thereby influencing the \textit{diversity} of queried samples and the overall clustering quality.
	Experiments are conducted on the UCM and AID datasets under the HAL configuration with $\lambda \in \{1.0,\ 2.0, \ 3.0, \ 4.0\}$.
	For each $\lambda$, the complete HAL loop is executed, with OA and AA reported across all annotation rounds to evaluate how $\lambda$ affects spectral clustering behavior and sample diversity.
	Under different values of $\lambda$, the quantitative results are summarized in Table~\ref{tab:hyperparamter_lambda}, while Fig.~\ref{fig:cluster_size_distribution} visualizes the corresponding distributions of spectral cluster sizes based on 2\% of the AID dataset.
	
	\begin{table*}[h]
		\center
		\caption{HAL’s performance under different values of $\lambda$ on UCM and AID datasets, reporting overall accuracy (OA) and average accuracy (AA). The best OA and AA at each labeling ratio (except 1\%) are highlighted in \textbf{bold}. }
		\label{tab:hyperparamter_lambda}	
		\setlength{\extrarowheight}{0mm}
		\setlength\tabcolsep{11pt}
		\resizebox{1.00\textwidth}{!}{
			\begin{tabular}{c | c | c c | c c| c c | c c| c c | c c}
				\hline			
				\hline		
				\multirow{2}*{Dataset}
				& \multirow{2}*{$\lambda$}
				& \multicolumn{2}{c|}{1\%}  
				& \multicolumn{2}{c|}{2\%}  
				& \multicolumn{2}{c|}{4\%}  
				& \multicolumn{2}{c|}{6\%}  
				& \multicolumn{2}{c|}{8\%}  
				& \multicolumn{2}{c}{10\%} \\ 
				\cline{3-14}
				
				& 
				& OA & AA & OA & AA  
				& OA & AA & OA & AA  
				& OA & AA & OA & AA  \\
				\hline
				
				\multirow{4}*{UCM} 
				& 1.0 
				& 54.8  & 52.7  & 57.4  & 57.6
				& 76.0  & 75.7  & 83.8  & 83.2
				& 86.0  & 85.5  & 93.3  & 92.8 \\ 
				
				& 2.0 
				& 56.4  & 55.8  & \textbf{63.3}  & \textbf{63.3}
				& 74.5  & 73.8  & 84.3  & 83.7
				& 90.7  & 90.3  & \textbf{94.5}  & \textbf{94.2} \\
				
				& 3.0 
				& 56.2  & 55.0  & 58.8  & 59.3
				& 77.9  & 77.7  & \textbf{87.1}  & \textbf{86.9}
				& 90.2  & 90.0  & 93.8  & 93.8 \\
				
				& 4.0 
				& 52.1  & 51.2  & 57.4  & 56.1
				& \textbf{78.3}  & \textbf{77.3}  & 85.2  & 84.5
				& \textbf{92.6}  & \textbf{92.2}  & 91.9  & 91.5 \\
				\hline
				
				\multirow{4}*{AID} 
				& 1.0 
				& 57.4  & 55.2  & 75.1  & 73.4
				& 85.7  & 84.5  & 90.1  & 89.6
				& 91.7  & 91.4  &94.3   & 94.1 \\ 
				
				& 2.0 
				& 58.7  & 56.0  & \textbf{75.9}  & \textbf{75.1}
				& 86.8  & 85.9  & 88.5  & 88.4
				& 93.0  & 92.8  & 94.1  & 93.7 \\
				
				& 3.0 
				& 54.2  & 51.7  & 74.0  & 75.5
				& 86.2  & 86.0  & \textbf{90.6}  & \textbf{90.4}
				& \textbf{93.8} & \textbf{93.4}  & \textbf{94.4} & \textbf{94.2}  \\
				
				& 4.0 
				& 54.2  & 51.6  & 72.1  & 71.5
				& \textbf{88.4}  & \textbf{87.8}  & 90.6  & 90.1
				& 92.2  & 92.1  & 94.2  & 94.0  \\
				
				\hline \hline
			\end{tabular}
		}
	\end{table*}

	As shown in Table~\ref{tab:hyperparamter_lambda}, varying the cluster multiplier $\lambda$ affects the granularity of spectral clustering and thereby the diversity of the queried samples. When $\lambda$ increases from 1.0 to 3.0, both OA and AA generally improve across UCM and AID, indicating that moderate cluster expansion enhances the representation of local structures and improves sample diversity. However, an excessively large $\lambda$ (e.g., 4.0) tends to over-fragment the clusters, as also reflected in Fig.~\ref{fig:cluster_size_distribution}, leading to unstable clustering and slight performance degradation.
	
	Overall, $\lambda=3.0$ achieves the best balance between stability and performance across both datasets and is therefore adopted as the default configuration for HAL. This setting effectively enhances the \textit{\textbf{diversity}} of queried samples by promoting finer yet coherent spectral clusters, enabling HAL to capture representative structures across the data manifold while maintaining strong and stable accuracy.
	
	\subsubsection{In-Cluster Selection Criterion (Uncertainty)}
	To examine the role of HAL's uncertainty in guiding in-cluster sample selection, we compare three strategies on UCM and AID:
	(i) \textit{\textbf{S-Random}} selection, which randomly picks one sample per cluster;
	(ii) \textit{\textbf{S-Centroid}} selection, which chooses the sample closest to the cluster center, representing the most representative instance; and
	(iii) \textit{\textbf{S-Uncertainty}} selection, which selects the sample with the highest uncertainty, measured by the gradient norm as described in Sec. \ref{sec::uncertainty_sampling}, to focus on ambiguous or hard-to-classify regions.
	All other settings of \textbf{HAL} remain the same. This comparison allows us to assess whether incorporating uncertainty helps the model identify more informative samples beyond those merely representative of the cluster structure.
	
	As shown in Table~\ref{tab:hyperparameter_incluster}, the \textit{\textbf{S-Centroid}} strategy already performs better than the \textit{\textbf{S-Random}} strategy, as selecting samples closer to cluster centers yields more representative and stable supervision.
	However, our \textit{\textbf{S-Uncertainty}} strategy further improves both OA and AA across all annotation ratios on UCM and AID in general. It achieves higher OA, indicating superior overall classification performance, and consistently higher AA, reflecting better class balance among the queried samples.
	This demonstrates that uncertainty-guided sampling enables HAL to focus on informative and ambiguous regions near decision boundaries while maintaining representative coverage, thereby improving both efficiency and fairness in sample selection.
	
	\begin{table*}[h]
		\center
		\caption{HAL’s performance of different sample selection strategies on UCM and AID datasets, reporting overall accuracy (OA) and average accuracy (AA). The best OA and AA at each labeling ratio (except 1\%) are highlighted in \textbf{bold}. }
		\label{tab:hyperparameter_incluster}	
		\setlength{\extrarowheight}{0mm}
		\setlength\tabcolsep{10pt}
		\resizebox{1.00\textwidth}{!}{
			\begin{tabular}{c | c | c c | c c| c c | c c| c c | c c}
				\hline			
				\hline		
				\multirow{2}*{Dataset}
				& {In-cluster}
				& \multicolumn{2}{c|}{1\%}  
				& \multicolumn{2}{c|}{2\%}  
				& \multicolumn{2}{c|}{4\%}  
				& \multicolumn{2}{c|}{6\%}  
				& \multicolumn{2}{c|}{8\%}  
				& \multicolumn{2}{c}{10\%} \\ 
				\cline{3-14}
				
				& Selection
				& OA & AA & OA & AA  
				& OA & AA & OA & AA  
				& OA & AA & OA & AA  \\
				\hline
				
				\multirow{3}*{UCM} 
				& {\textit{S-Random}} 
				& 56.1  & 56.3  & 57.6  & 57.6
				& 73.8  & 73.1  & 79.5  & 78.2
				& 86.9  & 86.6  & 87.6  & 87.3 \\ 
				
				& {\textit{S-Centroid}} 
				& 54.3  & 54.3  & 57.9  & 58.6
				& 74.5  & 74.0  & 84.0  & 83.4
				& 89.0  & 88.5  & 90.0  & 89.3 \\
				
				& {\textit{S-Uncertainty}} 
				& 56.2  & 55.0  & \textbf{58.8}  & \textbf{59.3}
				& \textbf{77.9}  & \textbf{77.7}  & \textbf{87.1}  & \textbf{86.9}
				& \textbf{90.2}  & \textbf{90.0}  & \textbf{93.8}  & \textbf{93.8} \\
				\hline
				
				\multirow{3}*{AID} 
				& {\textit{S-Random}} 
				& 57.3  & 54.1  & 72.5  & 72.4
				& 84.4  & 83.6  & 89.5  & 89.2
				& 91.3  & 90.9  & 94.2  & 93.9 \\ 
				
				& {\textit{S-Centroid}} 
				& 56.1  & 53.4  & \textbf{74.6}  & 73.0
				& \textbf{86.9}  & \textbf{86.3}  & 90.6  & 90.2
				& 91.7  & 91.2  & 92.4  & 92.2 \\
				
				& {\textit{S-Uncertainty}} 
				& 54.2  & 51.7  & 74.0  & \textbf{75.5}
				& 86.2  & 86.0  & \textbf{90.6}  & \textbf{90.4}
				& \textbf{93.8} & \textbf{93.4}  & \textbf{94.4} & \textbf{94.2}  \\
				
				\hline \hline
			\end{tabular}
		}
	\end{table*} 	
	
	\begin{table*}[h]
		\center
		\caption{Ablation study of HSSAL on UCM and AID datasets, reporting overall accuracy (OA) and average accuracy (AA). The best OA and AA at each labeling ratio (except 1\%) are highlighted in \textbf{bold}. }
		\label{tab:ablation_study}	
		\setlength{\extrarowheight}{0mm}
		\setlength\tabcolsep{8pt}
		\resizebox{1.00\textwidth}{!}{
			\begin{tabular}{c | c | c c | c c| c c | c c| c c | c c}
				\hline			
				\hline		
				\multirow{2}*{Dataset}
				& \multirow{2}*{Method}
				& \multicolumn{2}{c|}{1\%}  
				& \multicolumn{2}{c|}{2\%}  
				& \multicolumn{2}{c|}{4\%}  
				& \multicolumn{2}{c|}{6\%}  
				& \multicolumn{2}{c|}{8\%}  
				& \multicolumn{2}{c}{10\%} \\ 
				\cline{3-14}
				
				& 
				& OA & AA & OA & AA  
				& OA & AA & OA & AA  
				& OA & AA & OA & AA \\
				\hline
				
				\multirow{4}*{UCM} 
				& {Random} 
				& 57.3  & 56.4  & 59.5  & 58.1
				& 69.5  & 69.5  & 78.6  & 77.9 
				& 86.4  & 85.8  & 87.4  & 87.3 \\ 
				
				& {HAL} 
				& 56.2  & 55.0  & 58.8  & 59.3
				& 77.9  & 77.7  & 87.1  & 86.9
				& 90.2  & 90.0  & 93.8  & 93.8 \\
				
				& {Random + FixMatch} 
				& 52.6  & 51.8  & 56.2  & 55.2 
				& 65.7  & 64.0  & 81.9  & 80.9 
				& 85.7  & 85.0  & 88.6  & 87.3 \\ 
				
				& {HAL + FixMatch} 
				& 51.9  & 50.6  & \textbf{61.0}  & \textbf{60.3} 
				& \textbf{78.6}  & \textbf{78.6}  & \textbf{83.3}  & \textbf{82.6} 
				& \textbf{94.8}  & \textbf{94.5}  & \textbf{96.0}  & \textbf{95.9} \\ 
				\hline
				
				\multirow{4}*{AID} 
				& {Random} 
				& 57.3  & 54.1  & 72.5  & 72.4
				& 84.4  & 83.6  & 89.5  & 89.2
				& 91.3  & 90.9  & 94.2  & 93.9 \\ 
				
				& {HAL} 
				& 54.2  & 51.7  & 74.0  & 75.5
				& 86.2  & 86.0  & 90.6  & 90.4
				& 93.8  & 93.4  & 94.4  & 94.2  \\	
				
				& {Random + FixMatch} 
				& 56.5  & 53.6  & 78.3  & 76.9 
				& 87.9  & 86.8  & 93.5  & 93.0
				& 93.5  & 93.2  & 94.5  & 94.2 \\ 
				
				& {HAL + FixMatch} 
				& 61.3  & 57.7  & \textbf{83.2}  & \textbf{83.1} 
				& \textbf{91.5}  & \textbf{91.1}  & \textbf{95.2}  & \textbf{94.8} 
				& \textbf{95.1}  & \textbf{94.7}  & \textbf{96.4}  & \textbf{96.1} \\ 
				
				\hline \hline
			\end{tabular}
		}
	\end{table*}

	\begin{figure}[!t]
		\centering
		\begin{minipage}[b]{0.47\textwidth}
			\centering
			\includegraphics[width=\textwidth]{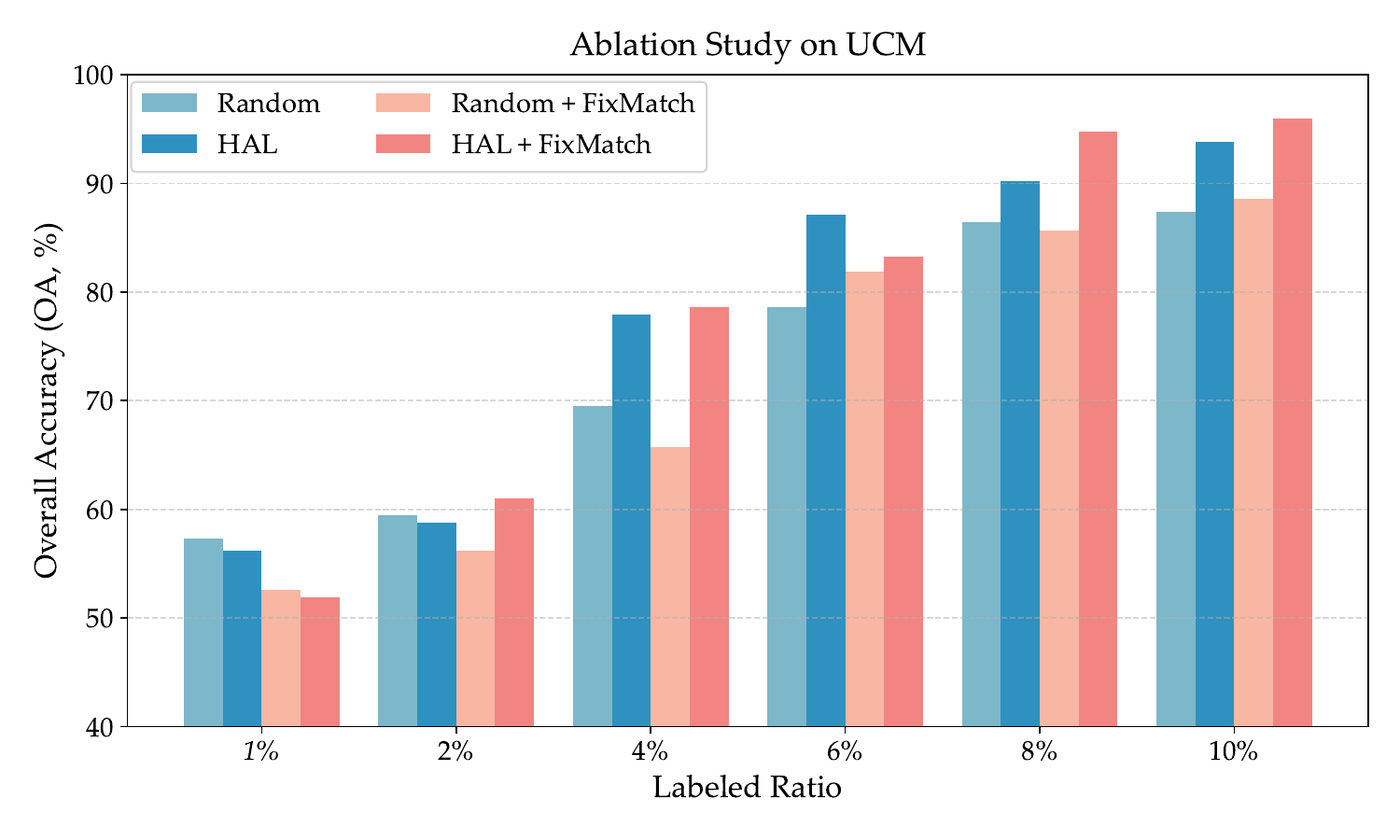}
			\label{fig:ablation_study_ucm}
			\vspace{-0.4cm}
		\end{minipage}
		\vspace{-0.5cm}
		\begin{minipage}[b]{0.47\textwidth}
			\centering
			\includegraphics[width=\textwidth]{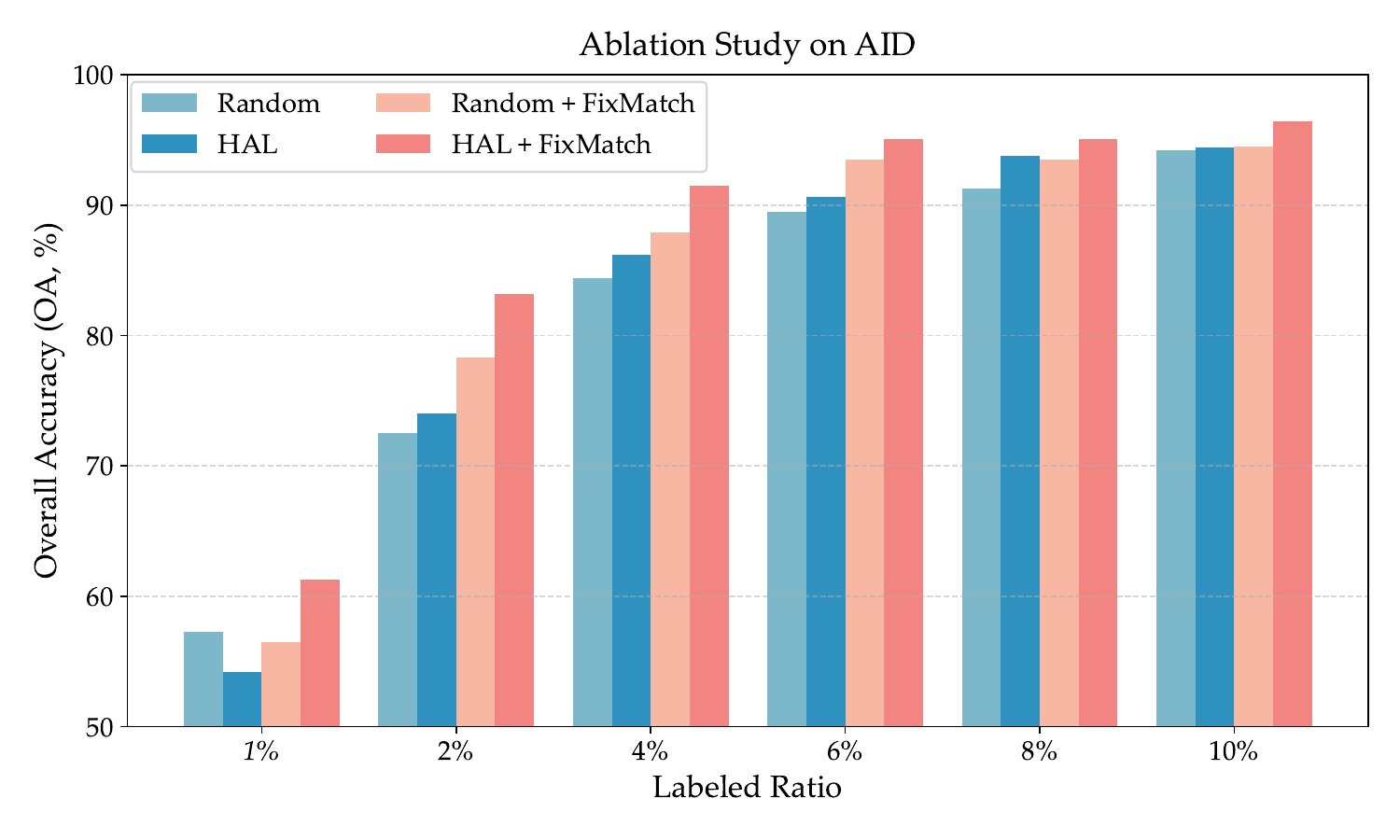}
			\label{fig:ablation_study_aid}
		\end{minipage}
		\caption{Performance comparison of HSSAL variants on UCM and AID datasets across different labeling ratios.}
		\label{fig:ablation_study}
	\end{figure}

    \begin{figure*}[!t]
        \centering
        \subfloat[Unlabeled Sample Distribution (2\% Random)\label{fig:ablation_study_true_distribution}]{
            \begin{minipage}[b]{0.48\textwidth}
                \centering
                \includegraphics[width=\textwidth]{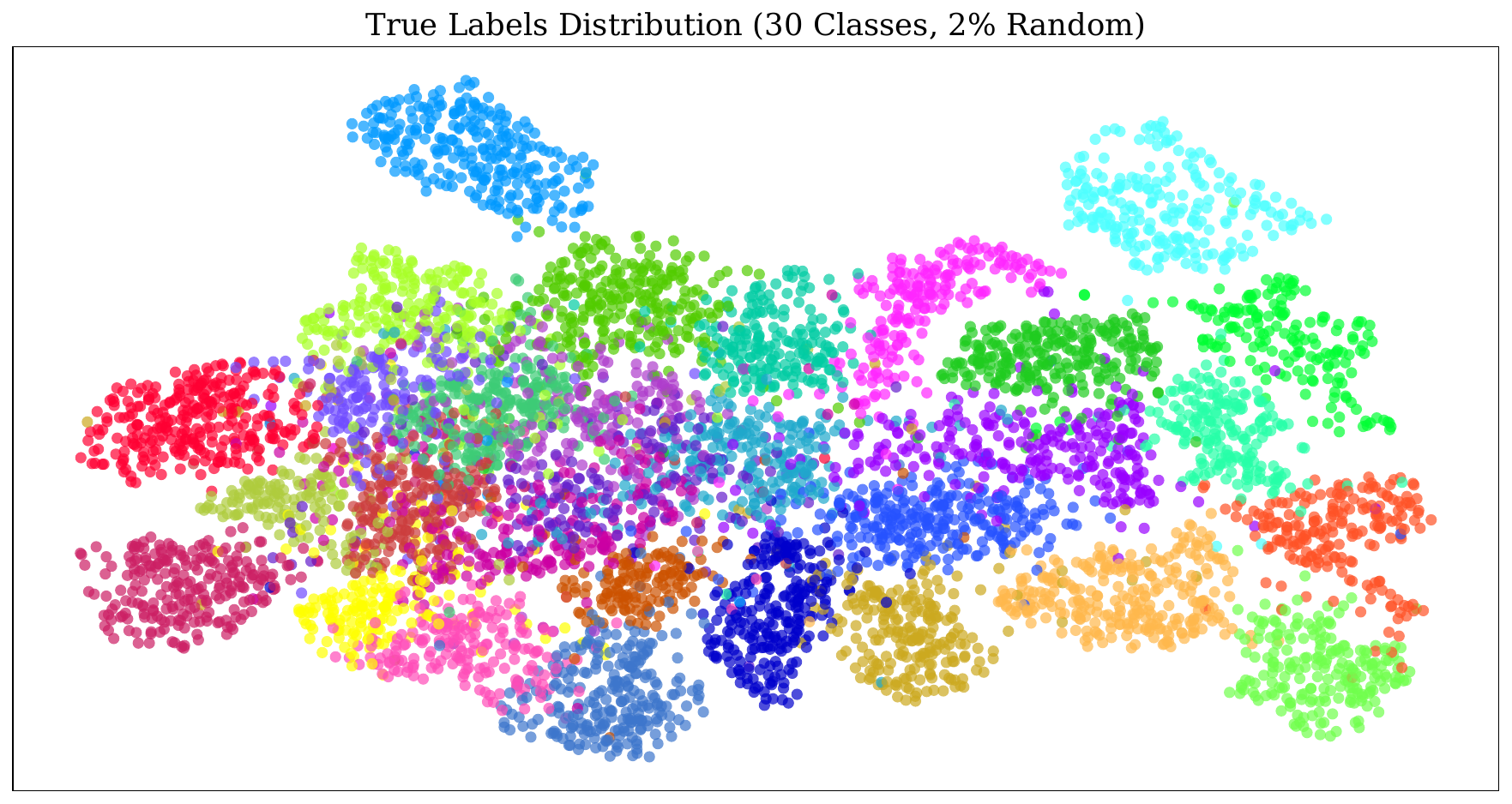}
            \end{minipage}
        }
        \hfill
        \subfloat[Unlabeled Sample Spectral Clustering (2\% FixMatch)\label{fig:ablation_study_spectral_clustering}]{
            \begin{minipage}[b]{0.48\textwidth}
                \centering
                \includegraphics[width=\textwidth]{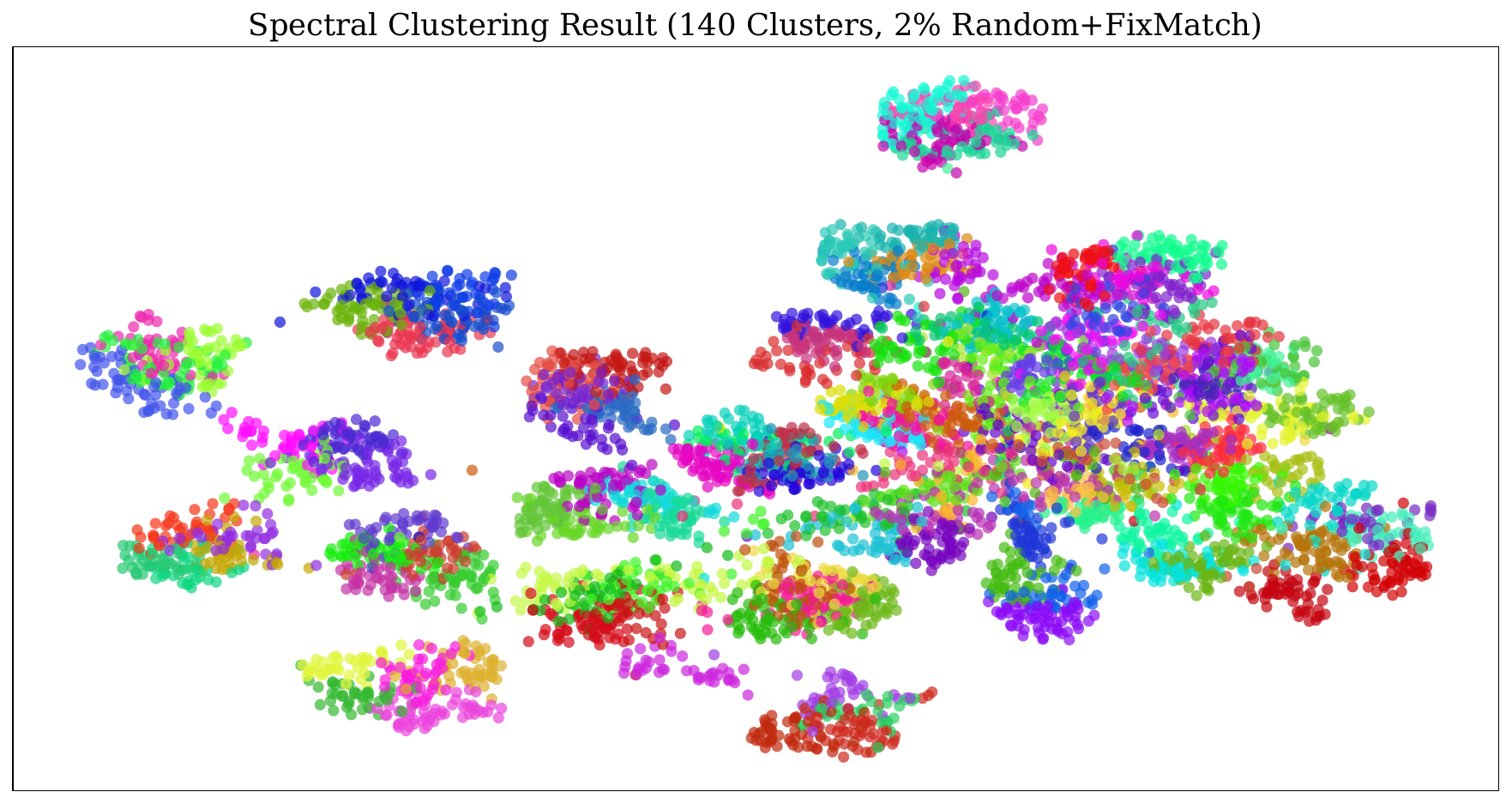}
            \end{minipage}
        }
        
        \vspace{0.2cm}
        
        \subfloat[Unlabeled Samples Selected by Random (2\% FixMatch)\label{fig:ablation_study_random_selection}]{
            \begin{minipage}[b]{0.48\textwidth}
                \centering
                \includegraphics[width=\textwidth]{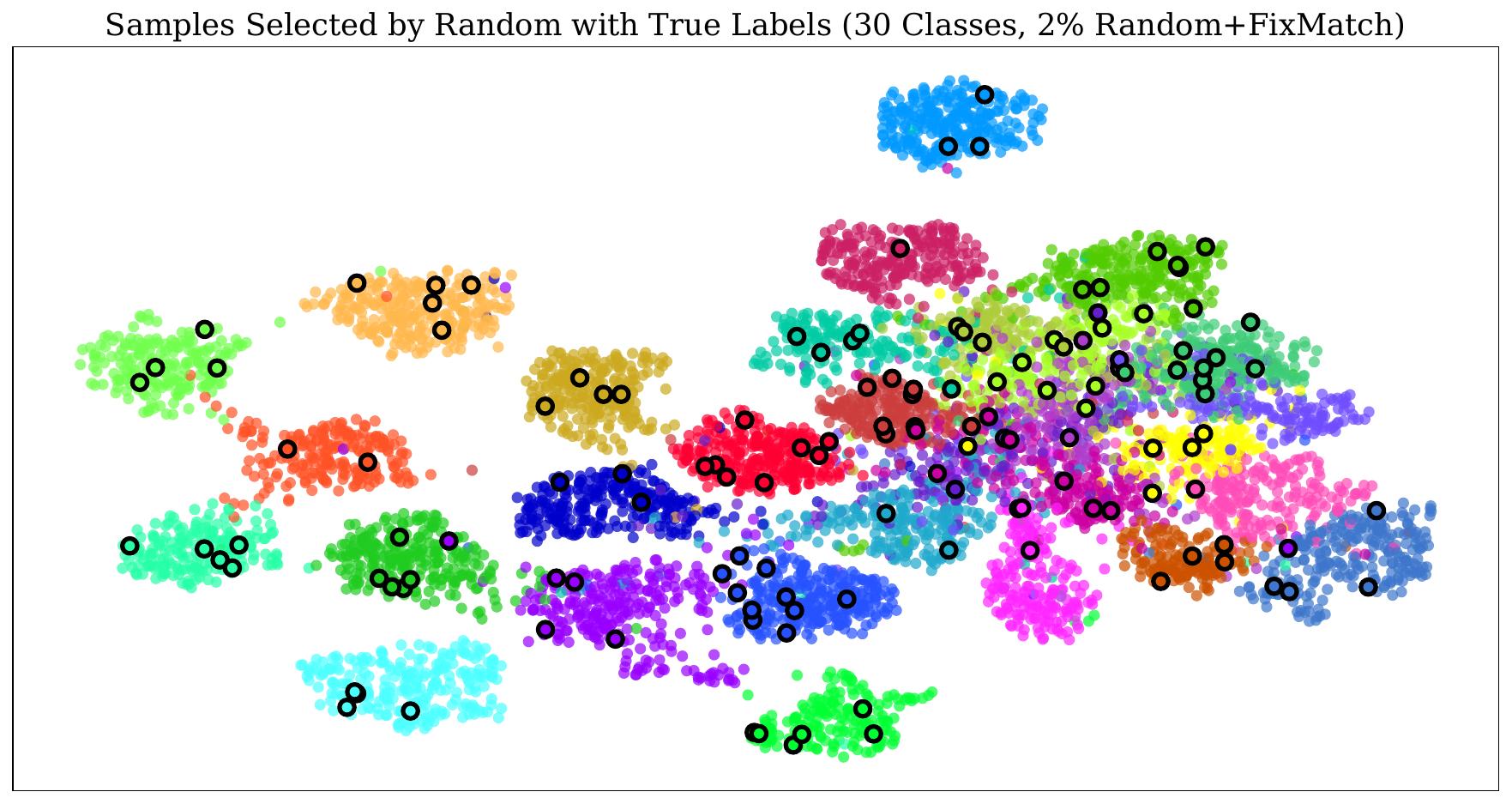}
            \end{minipage}
        }
        \hfill
        \subfloat[Unlabeled Samples Selected by HAL (2\% FixMatch)\label{fig:ablation_study_HAL_selection}]{
            \begin{minipage}[b]{0.48\textwidth}
                \centering
                \includegraphics[width=\textwidth]{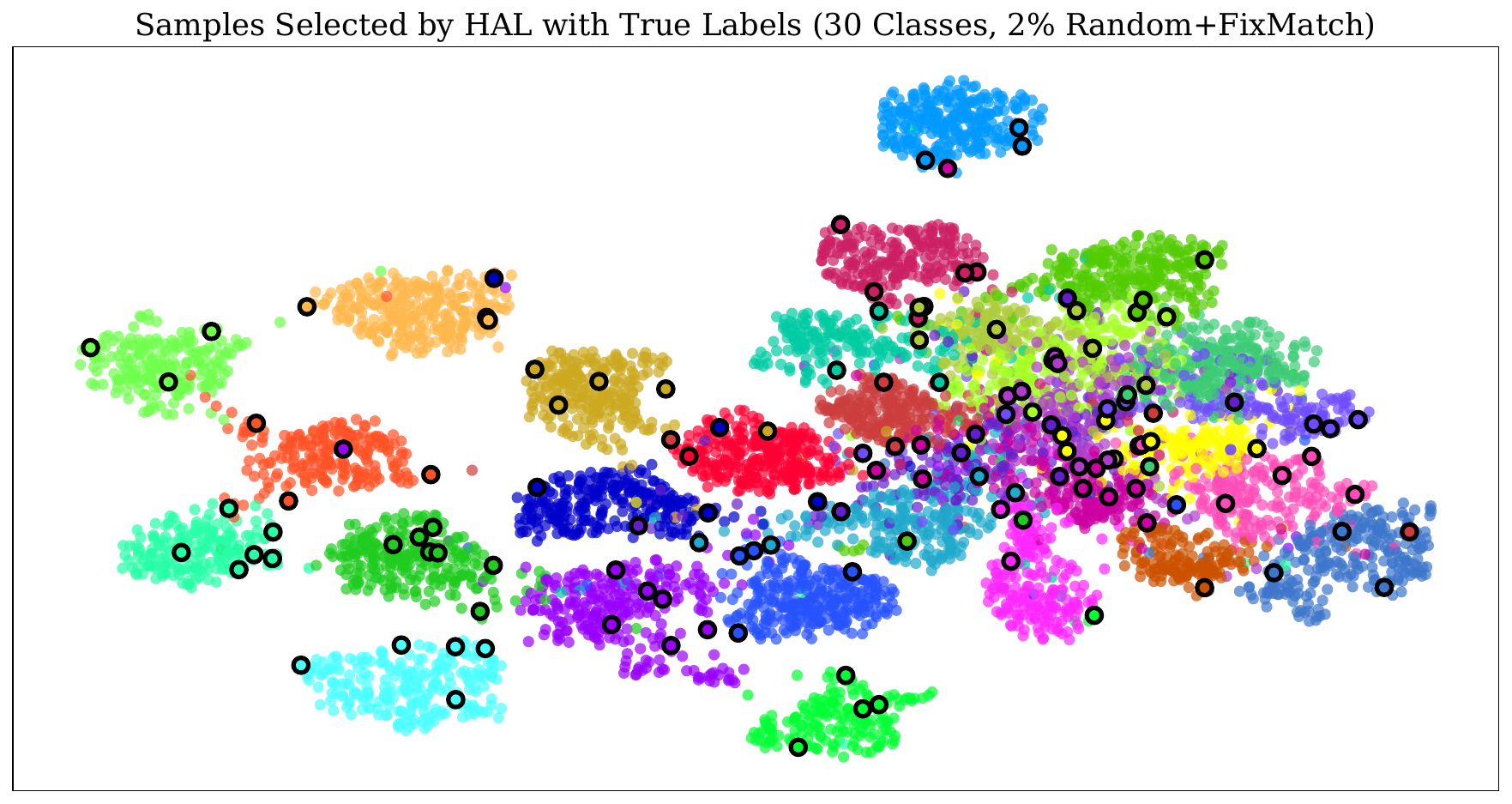}
            \end{minipage}
        }
        
        \caption{
            Distribution visualization of unlabeled training samples under the 2\% labeled setting on the AID dataset using the UMAP technique. 
            (a) True distribution of unlabeled samples based on  \textbf{Random}-trained model; 
            (b) Spectral clustering results with 140 clusters (for clarity, only 140 clusters are shown instead of 420 corresponding to $\lambda=3$); 
            (c) Unlabeled samples selected by the \textbf{Random} strategy based on \textbf{Random+FixMatch}-trained model; 
            (d) Unlabeled samples selected by the \textbf{HAL} strategy based on \textbf{Random+FixMatch}-trained model.
        }
        
        \label{fig:ablation_study_selection}
    \end{figure*}

	\subsection{Ablation study of HSSAL}
	\label{subsec:ablation}
	We perform an ablation to isolate the contributions of each component of HSSAL. Across the entire iterative training loops, we compare the following variants:
	
	\begin{itemize}
		\item \textbf{Random}: Random querying under the same budget on top of \emph{supervised-only} learning on labeled data. This provides a budget-controlled lower bound.
		\item \textbf{HAL}: Sample selection based on HAL on top of \emph{supervised-only} learning on labeled data.
		\item \textbf{Random+FixMatch}: The classical SSL baseline, FixMatch, that combines supervised learning on labeled data with self-training on unlabeled samples using a confidence threshold of $\tau=0.95$, \emph{\textbf{without (w/o)}} the HAL component.
		\item \textbf{HAL+FixMatch (HSSAL)}: The complete HSSAL framework that integrates FixMatch-based SSL with the proposed HAL for iterative model training and sample selection.
	\end{itemize}
	
	As shown in Table~\ref{tab:ablation_study} and Fig. \ref{fig:ablation_study}, the ablation results reveal that the individual contributions of SSL and AL differ notably between datasets of different scales. On the small-scale UCM dataset, introducing FixMatch (\textbf{Random+FixMatch}) for self-training on unlabeled data does not consistently improve performance compared with the supervised-only baseline (\textbf{Random}); in fact, it even degrades the OA under low labeling ratios, likely due to overfitting to noisy pseudo-labels caused by limited data diversity. In contrast, the \textbf{HAL} querying strategy shows clear gains over random selection, indicating its ability to identify more informative and representative samples. On the larger and more diverse AID dataset, the trend reverses: FixMatch provides significant improvements over random sampling, while the pure HAL variant contributes less prominently. This suggests that FixMatch benefits more from abundant unlabeled data, whereas HAL’s effectiveness is constrained when sample diversity is already high.
	
	When SSL and AL are combined in \textbf{HAL+FixMatch}, the performance consistently surpasses all other variants across both datasets. The synergy between HAL and FixMatch allows the framework to leverage SSL’s ability to reduce prediction uncertainty and AL’s capacity to select diverse and representative samples. This complementary interaction enhances model generalization and mitigates the instability observed when either component operates alone. Consequently, \textbf{HAL+FixMatch} achieves the most robust and scalable performance, demonstrating the coordinated role of SSL and AL in jointly improving uncertainty reduction and feature representation.

	Furthermore, the visualization results of unlabeled training samples under the 2\% labeled setting based on the models trained by \textbf{Random} and \textbf{Random+FixMatch} are illustrated in Fig.~\ref{fig:ablation_study_selection}, using the UMAP technique\cite{mcinnes2018umap}. Fig.~\ref{fig:ablation_study_true_distribution} presents the true distribution of the \textbf{Random} strategy, while Figs.~\ref{fig:ablation_study_random_selection} and~\ref{fig:ablation_study_HAL_selection} visualize the unlabeled samples selected by \textbf{Random} and \textbf{HAL}, respectively. In addition, Fig.~\ref{fig:ablation_study_spectral_clustering} displays the spectral clustering results with 140 clusters (for clarity, only 140 clusters are shown instead of 420 corresponding to $\lambda = 3$), where the clustering process partitions the dataset into multiple locally coherent subsets, providing an effective foundation for HAL’s subsequent informative sample selection shown in Fig.~\ref{fig:ablation_study_HAL_selection}.
	The comparison between Fig.~\ref{fig:ablation_study_true_distribution} and Figs.~\ref{fig:ablation_study_random_selection}–\ref{fig:ablation_study_HAL_selection} indicates that SSL can effectively enhance the discrimination of class boundaries. Moreover, by comparing Figs.~\ref{fig:ablation_study_random_selection} and~\ref{fig:ablation_study_HAL_selection}, it can be observed that HAL tends to select more informative samples that are both diverse across the entire dataset and highly uncertain near the class boundaries.
	
	\subsection{Comparison experiments}
	\label{Experiments}
	To comprehensively evaluate the effectiveness of the proposed \textbf{HSSAL} framework, we compare it with six representative \textbf{AL} strategies and four state-of-the-art \textbf{SSL} methods.
	The AL baselines encompass both uncertainty- and diversity-driven querying strategies, while the SSL baselines represent different pseudo-labeling paradigms commonly adopted in recent works.
	Furthermore, we examine the combined variants (\textbf{HAL+SSL}, \textit{i.e.}, \textbf{HSSAL}) to assess the complementarity and robustness of our hierarchical active learning strategy when integrated with different SSL algorithms under limited annotation scenarios.

	\begin{table*}[h]
		\center
		\caption{Comparison results on the UCM dataset among \textbf{HAL}, various \textbf{AL} strategies, \textbf{SSL} methods, and their integrated variants (\textbf{HSSAL}). The table reports overall accuracy (OA) and average accuracy (AA) under different labeling ratios. The best OA and AA at each ratio (except 1\%) and their average values are highlighted in \textbf{bold}.}
		\label{tab:comparison_ucm}	
		\setlength{\extrarowheight}{0mm}
		\setlength\tabcolsep{6pt}
		\resizebox{1.00\textwidth}{!}{
			\begin{tabular}{c | c | c c | c c| c c | c c| c c | c c | c c }
				\hline			
				\hline		
				\multirow{2}*{Type}
				& \multirow{2}*{Method}
				& \multicolumn{2}{c|}{1\%}  
				& \multicolumn{2}{c|}{2\%}  
				& \multicolumn{2}{c|}{4\%}  
				& \multicolumn{2}{c|}{6\%}  
				& \multicolumn{2}{c|}{8\%}  
				& \multicolumn{2}{c|}{10\%} 
				& \multicolumn{2}{c}{Avg (2\%-10\%)} \\ 
				\cline{3-16}
				
				& 
				& OA & AA & OA & AA  
				& OA & AA & OA & AA  
				& OA & AA & OA & AA
				& OA & AA  \\
				\hline
				
				\multirow{6}*{AL} 
				& {Random} 
				& 57.3  & 56.4  & 59.5  & 58.1
				& 69.5  & 69.5  & 78.6  & 77.9 
				& 86.4  & 85.8  & 87.4  & 87.3
				& 76.3  & 75.7  \\ 
				
				& {Margin} 
				& 54.1  & 53.2  & 47.1  & 46.4  
				& 51.4  & 50.6  & 52.6  & 51.8  
				& 54.0  & 52.7  & 55.0  & 54.2  
				& 52.0  & 51.1  \\ 
				
				& {Entropy} 
				& 55.7  & 56.4  & 58.6  & 57.9  
				& 59.3  & 57.9  & 67.9  & 68.6  
				& 76.4  & 76.5  & 86.0  & 86.2 
				& 69.6  & 69.4  \\ 
				
				& {BALD} 
				& 55.0  & 55.0  & 51.0  & 50.5  
				& 67.4  & 65.7  & 81.7  & 80.3  
				& 82.4  & 83.0  & 89.3  & 88.9  
				& 74.4  & 73.7  \\ 
				
				& {CoreSet} 
				& 56.7  & 56.0  & 58.1  & 56.2  
				& 67.6  & 66.4  & 85.5  & 84.8  
				& 89.0  & 88.4  & 92.1  & 92.0  
				& 78.5  & 77.6  \\ 
				
				& {BADGE} 
				& 52.4  & 51.4  & 57.9  & 56.8  
				& 72.1  & 71.3  & 81.4  & 80.2  
				& 90.0  & 89.1  & \textbf{94.8}  & \textbf{94.6}  
				& 79.2  & 78.4  \\ 
				
				& {\textbf{\textit{Our HAL}}} 
				& 56.2  & 55.0  & \textbf{58.8}  & \textbf{59.3}
				& \textbf{77.9}  & \textbf{77.7}  & \textbf{87.1}  & \textbf{86.9}
				& \textbf{90.2}  & \textbf{90.0}  & 93.8  & 93.8
				& \textbf{81.6}  & \textbf{81.5}  \\ 
				\hline

				& {Random+FixMatch} 
				& 52.6  & 51.8  & 56.2  & 55.2 
				& 65.7  & 64.0  & 81.9  & 80.9 
				& 85.7  & 85.0  & 88.6  & 87.3
				& 75.6  & 74.5  \\ 
				
				& {\textit{\textbf{HAL+FixMatch}}} 
				& 51.9  & 50.6  & \textbf{61.0}  & \textbf{60.3} 
				& \textbf{78.6}  & \textbf{78.6}  & \textbf{83.3}  & \textbf{82.6} 
				& \textbf{94.8}  & \textbf{94.5}  & \textbf{96.0}  & \textbf{95.9}
				& \textbf{82.7}  & \textbf{82.4}  \\ 
				\cline{2-16}
				
				& {Random+FlexMatch} 
				& 47.6  & 46.2  & 62.4  & 62.6  
				& 73.6  & 73.9  & 86.2  & 85.6  
				& 90.2  & 90.0  & 90.2  & 89.9  
				& 80.5  & 804  \\ 
				
				SSL/        
				& {\textit{\textbf{HAL+FlexMatch}}} 
				& 36.9  & 37.1  & \textbf{75.5}  & \textbf{74.4}  
				& \textbf{87.6}  & \textbf{87.4}  & \textbf{87.6}  & \textbf{87.2}  
				& \textbf{94.5}  & \textbf{94.4}  & \textbf{95.5}  & \textbf{95.5} 
				& \textbf{88.1}  & \textbf{87.8}  \\ 
				
				\cline{2-16}
				
				HSSAL
				& {Random+FreeMatch} 
				& 57.9  & 58.1  & 56.7  & 54.6  
				& 68.1  & 67.9  & 75.5  & 75.7  
				& 79.8  & 80.2  & 84.8  & 84.7  
				& 73.0  & 72.6  \\ 
				
				& {\textit{\textbf{HAL+FreeMatch}}} 
				& 58.3  & 57.4  & \textbf{59.5}  & \textbf{58.1}  
				& \textbf{82.6}  & \textbf{82.8}  & \textbf{80.5}  & \textbf{80.2}  
				& \textbf{84.8}  & \textbf{84.5}  & \textbf{93.8}  & \textbf{93.1}  
				& \textbf{80.2}  & \textbf{79.7}  \\ 
				\cline{2-16}
				
				& {Random+SoftMatch} 
				& 52.6  & 52.7  & 5.0  & 4.8  
				& 5.0  & 4.8   & 5.0  & 4.8 
				& 5.0  & 4.8   & 5.0  & 4.8
				& 5.0  & 4.8  \\ 
				
				& {\textit{\textbf{HAL+SoftMatch}}} 
				& 51.4  & 50.7  & \textbf{60.5}  & \textbf{58.9}  
				& \textbf{79.3}  & \textbf{78.2}  & \textbf{80.2}  & \textbf{79.6}  
				& \textbf{87.9}  & \textbf{87.4}  & \textbf{88.6}  & \textbf{88.0} 
				& \textbf{79.3}  & \textbf{78.4}  \\ 
				\hline
				
				\multicolumn{2}{c|}{Fully-supervised Learning (Oracle)}   
				& \multicolumn{14}{c}{100\%: OA: 99.1, AA: 99.1} \\
				\hline \hline
			\end{tabular}
		}
	\end{table*}

	\begin{table*}[h]
		\center
		\caption{Comparison results on the AID dataset among \textbf{HAL}, various \textbf{AL} strategies, \textbf{SSL} methods, and their integrated variants (\textbf{HSSAL}). The table reports overall accuracy (OA) and average accuracy (AA) under different labeling ratios. The best OA and AA at each ratio (except 1\%) and their average values are highlighted in \textbf{bold}.}
		\label{tab:comparison_aid}	
		\setlength{\extrarowheight}{0mm}
		\setlength\tabcolsep{6pt}
		\resizebox{1.00\textwidth}{!}{
			\begin{tabular}{c | c | c c | c c| c c | c c| c c | c c | c c }
				\hline			
				\hline		
				\multirow{2}*{Type}
				& \multirow{2}*{Method}
				& \multicolumn{2}{c|}{1\%}  
				& \multicolumn{2}{c|}{2\%}  
				& \multicolumn{2}{c|}{4\%}  
				& \multicolumn{2}{c|}{6\%}  
				& \multicolumn{2}{c|}{8\%}  
				& \multicolumn{2}{c|}{10\%} 
				& \multicolumn{2}{c}{Avg (2\%-10\%)} \\ 
				\cline{3-16}
				
				& 
				& OA & AA & OA & AA  
				& OA & AA & OA & AA  
				& OA & AA & OA & AA
				& OA & AA  \\
				\hline
				
				\multirow{6}*{AL} 
				& {Random} 
				& 57.3  & 54.1  & 72.5  & 72.4
				& 84.4  & 83.6  & 89.5  & 89.2
				& 91.3  & 90.9  & 94.2  & 93.9
				& 86.4  & 86.0  \\ 
				
				& {Margin} 
				& 51.9  & 49.1  & 52.2  & 50.8  
				& 53.0  & 50.0  & 55.7  & 53.1  
				& 58.5  & 55.4  & 60.0  & 56.9  
				& 55.9  & 53.2  \\ 
				
				& {Entropy} 
				& 57.4  & 54.2  & 67.5  & 66.8  
				& 78.9  & 78.8  & 87.8  & 87.4  
				& 90.5  & 90.2  & 93.9  & 93.7  
				& 83.7  & 83.4  \\ 
				
				& {BALD} 
				& 57.5  & 55.2  & 65.2  & 64.7  
				& 80.9  & 80.4  & 87.9  & 87.6  
				& 91.0  & 90.4  & 94.1  & 93.9 
				& 83.8  & 83.4  \\ 
				
				& {CoreSet} 
				& 54.2  & 51.5  & 72.1  & 70.8  
				& \textbf{86.3}  & 85.9  & 89.5  & 89.4 
				& 91.7  & 91.4  & 94.2  & 93.9  
				& 86.8  & 86.3  \\ 
				
				& {BADGE} 
				& 57.8  & 55.1  & 72.2  & 71.4  
				& 84.8  & 84.1  & 88.4  & 87.8  
				& 91.5  & 91.0  & 93.4  & 93.3  
				& 86.1  & 85.5  \\ 
				
				& {\textbf{\textit{Our HAL}}} 
				& 54.2  & 51.7  & \textbf{74.0}  & \textbf{75.5}
				& 86.2  & \textbf{86.0}  & \textbf{90.6}  & \textbf{90.4}
				& \textbf{93.8}  & \textbf{93.4}  & \textbf{94.4}  & \textbf{94.2}
				& \textbf{87.8}  & \textbf{87.9}  \\ 
				\hline

				& {Random+FixMatch} 
				& 56.5  & 53.6  & 78.3  & 76.9 
				& 87.9  & 86.8  & 93.5  & 93.0
				& 93.5  & 93.2  & 94.5  & 94.2
				& 89.5  & 88.8  \\ 
				
				& {\textit{\textbf{HAL+FixMatch}}} 
				& 61.3  & 57.7  & \textbf{83.2}  & \textbf{83.1} 
				& \textbf{91.5}  & \textbf{91.1}  & \textbf{95.2}  & \textbf{94.8} 
				& \textbf{95.1}  & \textbf{94.7}  & \textbf{96.4}  & \textbf{96.1} 
				& \textbf{92.3}  & \textbf{92.0}  \\ 
				\cline{2-16}
				
				& {Random+FlexMatch} 
				& 79.6  & 78.8  & 83.6  & 82.8  
				& 90.7  & 90.1  & 92.9  & 92.6  
				& 93.1  & 92.9  & 94.1  & 93.8  
				& 90.9  & 90.4  \\ 
				
				SSL/        
				& {\textit{\textbf{HAL+FlexMatch}}} 
				& 73.0  & 71.8  & \textbf{91.6}  & \textbf{91.4}  
				& \textbf{93.4}  & \textbf{93.2}  & \textbf{94.7}  & \textbf{94.5}  
				& \textbf{95.5}  & \textbf{95.1}  & \textbf{96.0}  & \textbf{95.8}   
				& \textbf{94.2}  & \textbf{94.0}  \\ 
				\cline{2-16}
				
				HSSAL
				& {Random+FreeMatch} 
				& 67.3  & 65.2  & 76.3  & 75.2  
				& 90.5  & 89.8  & 92.2  & 91.7  
				& 94.1  & 93.6  & 95.1  & 94.8  
				& 89.6  & 89.0  \\ 
				
				& {\textit{\textbf{HAL+FreeMatch}}} 
				& 73.3  & 68.9  & \textbf{79.4}  & \textbf{77.6}  
				& \textbf{92.8}  & \textbf{92.8}  & \textbf{95.1}  & \textbf{95.0}  
				& \textbf{96.1}  & \textbf{95.9}  & \textbf{96.3}  & \textbf{96.1}  
				& \textbf{91.9}  & \textbf{91.5}  \\ 
				\cline{2-16}
				
				& {Random+SoftMatch} 
				& 68.9  & 66.8  & 75.6  & 73.8  
				& 89.1  & 88.5  & 90.7  & 90.6  
				& 92.5  & 92.2  & 93.8  & 93.6  
				& 88.3  & 87.5  \\ 
				
				& {\textit{\textbf{HAL+SoftMatch}}} 
				& 65.2  & 64.7  & \textbf{86.3}  & \textbf{86.0}  
				& \textbf{91.4}  & \textbf{90.7}  & \textbf{94.8}  & \textbf{94.5}  
				& \textbf{95.0}  & \textbf{94.6}  & \textbf{95.9}  & \textbf{95.7}  
				& \textbf{92.7}  & \textbf{92.3}  \\ 
				\hline
				
				\multicolumn{2}{c|}{Fully-supervised Learning (Oracle)}   
				& \multicolumn{14}{c}{100\%: OA: 97.9, AA: 97.7}
				\\
				
				\hline \hline
				
			\end{tabular}
		}
	\end{table*}

	\begin{table*}[h]
		\center
		\caption{Comparison results on the NWPU-RESISC45 dataset among \textbf{HAL}, various \textbf{AL} strategies, \textbf{SSL} methods, and their integrated variants (\textbf{HSSAL}). The table reports overall accuracy (OA) and average accuracy (AA) under different labeling ratios. The best OA and AA at each ratio (except 1\%) and their average values are highlighted in \textbf{bold}.}
		\label{tab:comparison_nwpu_resisc45}	
		\setlength{\extrarowheight}{0mm}
		\setlength\tabcolsep{6pt}
		\resizebox{1.00\textwidth}{!}{
			\begin{tabular}{c | c | c c | c c| c c | c c| c c | c c | c c }
				\hline			
				\hline		
				\multirow{2}*{Type}
				& \multirow{2}*{Method}
				& \multicolumn{2}{c|}{1\%}  
				& \multicolumn{2}{c|}{2\%}  
				& \multicolumn{2}{c|}{4\%}  
				& \multicolumn{2}{c|}{6\%}  
				& \multicolumn{2}{c|}{8\%}  
				& \multicolumn{2}{c|}{10\%} 
				& \multicolumn{2}{c}{Avg (2\%-10\%)} \\ 
				\cline{3-16}
				
				& 
				& OA & AA & OA & AA  
				& OA & AA & OA & AA  
				& OA & AA & OA & AA
				& OA & AA  \\
				\hline
				
				\multirow{6}*{AL} 
				& {Random} 
				& 75.2  & 75.2  & 79.7  & 79.9  
				& 87.1  & 87.2  & 90.0  & 90.0  
				& 91.1  & 91.2  & 92.1  & 92.1  
				& 88.0  & 88.1  \\ 
				
				& {Margin} 
				& 71.4  & 71.6  & 71.0  & 71.2  
				& 73.0  & 73.1  & 70.0  & 70.1  
				& 76.7  & 76.9  & 72.1  & 72.3  
				& 72.6  & 72.7  \\ 
				
				& {Entropy} 
				& 72.2  & 72.1  & 78.7  & 78.7  
				& 89.0  & 89.1  & 91.3  & 91.3  
				& 93.0  & 93.1  & 94.6  & 94.6  
				& 89.3  & 89.4  \\ 
				
				& {BALD} 
				& 71.3  & 71.6  & 81.3  & 81.4  
				& 86.6  & 86.7  & 90.9  & 90.9  
				& 92.9  & 92.9  & \textbf{95.0}  & \textbf{95.0 } 
				& 89.3  & 89.4  \\ 
				
				& {CoreSet} 
				& 72.7  & 72.8  & 82.5  & 82.6  
				& 90.1  & 90.1  & 92.4  & 92.4  
				& 93.1  & 93.2  & \textbf{95.0}  & \textbf{95.0}  
				& 90.6  & 90.7  \\ 
				
				& {BADGE} 
				& 72.1  & 72.3  & 82.8  & 82.9  
				& 89.7  & 89.7  & 93.4  & 93.4  
				& \textbf{94.0}  & \textbf{93.9}  & 94.8  & 94.8 
				& 90.9  & 90.9  \\ 
				
				& {\textbf{\textit{Our HAL}}} 
				& 72.4  & 72.5  & \textbf{84.5}  & \textbf{84.4}   
				& \textbf{91.0}  & \textbf{91.1} & \textbf{93.4}  & \textbf{93.4} 
				& 93.7  & 93.8  & 94.7  & 94.7  				
				& \textbf{91.5}  & \textbf{91.5}  \\ 
				\hline
				
				& {Random+FixMatch} 
				& 88.5  & 88.5  & 90.2  & 90.2  
				& 93.2  & 93.2  & 93.9  & 93.9  
				& 94.3  & 94.3  & 94.7  & 94.7  
				& 93.3  & 93.3  \\ 
				
				& {\textit{\textbf{HAL+FixMatch}}} 
				& \textbf{86.3}  & \textbf{86.4}  & \textbf{90.3}  & \textbf{90.2}  
				& \textbf{94.4}  & \textbf{94.5}  & \textbf{95.8}  & \textbf{95.8}   
				& \textbf{95.3}  & \textbf{95.4}  & \textbf{96.5}  & \textbf{96.5}  
				& \textbf{94.5}  & \textbf{94.5}  \\ 
				\cline{2-16}
				
				& {Random+FlexMatch} 
				& 88.1  & 88.2  & 90.1  & 90.2  
				& 92.5  & 92.6  & 93.8  & 93.8  
				& 94.3  & 94.3  & 94.8  & 94.8  
				& 93.1  & 93.1  \\ 
				
				SSL/        
				& {\textit{\textbf{HAL+FlexMatch}}} 
				& \textbf{90.3}  & \textbf{90.3}  & \textbf{92.3}  & \textbf{92.4}  
				& \textbf{94.5}  & \textbf{94.5}  & \textbf{95.5}  & \textbf{95.5}  
				& \textbf{96.0}  & \textbf{96.0}  & \textbf{96.6}  & \textbf{96.6}  
				& \textbf{95.0}  & \textbf{95.0}  \\ 
				
				\cline{2-16}
				
				HSSAL
				& {Random+FreeMatch} 
				& 85.1  & 85.2  & 90.4  & 90.2  
				& 93.1  & 93.1  & 93.2  & 93.2  
				& 94.7  & 94.7  & 95.3  & 95.3  
				& 93.3  & 93.3  \\ 
				
				& {\textit{\textbf{HAL+FreeMatch}}} 
				& 82.4  & 82.7  & \textbf{90.9}  & \textbf{90.9}  
				& \textbf{94.7}  & \textbf{94.8}  & \textbf{94.4}  & \textbf{94.4}  
				& \textbf{95.0}  & \textbf{95.0}  & \textbf{96.6}  & \textbf{96.6}  
				& \textbf{94.3}  & \textbf{94.3}  \\ 
				\cline{2-16}
				
				& {Random+SoftMatch} 
				& 79.0  & 79.2  & 90.6  & 90.6  
				& 91.5  & 91.5  & 93.0  & 93.1   
				& 93.9  & 93.9  & 94.9  & 95.0  
				& 92.8  & 92.8  \\ 
				
				& {\textit{\textbf{HAL+SoftMatch}}} 
				& \textbf{80.8}  & \textbf{80.7}  & \textbf{91.4}  & \textbf{91.4}  
				& \textbf{93.7}  & \textbf{93.7}  & \textbf{94.5}  & \textbf{94.5}  
				& \textbf{94.9}  & \textbf{94.9}  & \textbf{96.0}  & \textbf{96.0}  
				& \textbf{94.1}  & \textbf{94.1}  \\ 
				\hline
				
				\multicolumn{2}{c|}{Fully-supervised Learning (Oracle)}   
				& \multicolumn{14}{c}{100\%: OA: 96.0, AA: 96.0} \\
				\hline \hline
			\end{tabular}
		}
	\end{table*} 	
	
	Specifically, the six AL baselines include:
	\begin{itemize}
		\item \textbf{Random}: Randomly selects samples for annotation under the same labeling budget as a baseline reference.
		\item \textbf{Entropy}\cite{settles2009active}: Queries samples with the highest predictive entropy, reflecting maximum model uncertainty.
		\item \textbf{Margin}\cite{scheffer2001active}: Selects samples with the smallest difference between the top two class probabilities, focusing on ambiguous predictions near the decision boundary.
		\item \textbf{BALD}\cite{bald}: Estimates both epistemic and aleatoric uncertainty by maximizing the mutual information between predictions and model parameters. We adopt Monte Carlo dropout with $T=50$ stochastic forward passes on the last layer to approximate the acquisition score.
		\item \textbf{CoreSet}\cite{coreset}: Selects representative samples via $k$-center clustering in the feature space. The image features are the last-layer \texttt{[CLS]} tokens from the DINOv2-Small encoder, identical to those used in HAL.		
		\item \textbf{BADGE}\cite{badge}: Combines uncertainty and diversity by clustering gradient embeddings with $k$-means++. It operates on the same DINOv2 \texttt{[CLS]} features as CoreSet and HAL.
	\end{itemize}
	
	In addition, the following SSL methods are adopted for comparison and integration with HAL:
	\begin{itemize}
		\item \textbf{FixMatch}\cite{sohn2020fixmatch}: A representative weak-to-strong consistency method that generates pseudo-labels from weakly augmented images and enforces consistency on their strongly augmented counterparts when confidence exceeds a fixed threshold $\tau=0.95$.
		\item \textbf{FlexMatch}\cite{zhang2021flexmatch}: Builds on FixMatch by introducing a class-wise adaptive threshold $\tau_c$ according to the learning progress of each class to reduce confirmation bias.
		\item \textbf{FreeMatch}\cite{wang2023freematch}: Further improves FixMatch by dynamically adjusting both global and class-specific thresholds based on prediction history for adaptive pseudo-labeling.
		\item \textbf{SoftMatch}\cite{chen2023softmatch}: Extends the FixMatch paradigm by replacing the hard threshold with Gaussian confidence weighting, enabling smoother exploitation of uncertain samples.
	\end{itemize}
	
	As shown in Tables~\ref{tab:comparison_ucm}–\ref{tab:comparison_nwpu_resisc45}, among all AL baselines, the proposed HAL consistently achieves the highest OA and AA across different labeling ratios. The close alignment between OA and AA indicates that HAL not only selects the most informative samples but also maintains class balance during the annotation process. This advantage mainly stems from its hierarchical sample selection mechanism, which ensures both representativeness and diversity in sample selection from the unlabeled pool.
	
	When integrated with different SSL algorithms, all HAL+SSL variants, referred to as HSSAL, achieve substantial improvements compared with their respective HAL and SSL counterparts. The results highlight the complementary effect of active querying and pseudo-label–based self-training, demonstrating that the two paradigms can jointly enhance learning efficiency when annotation resources are limited.
	
	The synergy between HAL and SSL arises from their distinct yet mutually reinforcing treatment of uncertainty. HAL focuses on discovering highly uncertain and representative samples for manual labeling, while SSL adaptively generates pseudo-labels for low-uncertainty samples through dynamic or class-wise thresholding strategies, such as those used in FlexMatch, FreeMatch, and SoftMatch. This collaborative mechanism enables the model to exploit both uncertain and confident samples effectively, thereby maximizing the overall utilization of the unlabeled data.
	
	Among all SSL algorithms, FlexMatch exhibits the most stable and robust performance. When combined with HAL, it achieves more than \textbf{95\%} of the fully supervised accuracy the NWPU-RESISC45, AID, and UCM datasets using only 2\%, 4\%, and 8\% labeled samples, respectively. The results demonstrate that the proposed HSSAL framework effectively integrates hierarchical active querying with adaptive semi-supervised learning, achieving superior label efficiency, robustness, and generalization under low-annotation conditions.

	\section{Conclusion}
	This paper proposed a unified \textbf{HSSAL} framework for remote sensing scene classification under limited annotation budgets. HSSAL integrates a weak-to-strong semi-supervised learning (SSL) paradigm with a hierarchical active learning (HAL) strategy in an iterative SSL–HAL loop, effectively exploiting unlabeled data across different uncertainty levels. In each round, SSL refines feature representations using both labeled and confidently pseudo-labeled samples, while HAL performs sample querying that jointly considers \textit{\textbf{scalability}}, \textit{\textbf{diversity}}, and \textit{\textbf{uncertainty}} to select the most informative instances for annotation.
	
	The proposed HAL jointly considers scalability, diversity, and uncertainty through mini-batch partitioning, spectral clustering, and gradient-based selection. This design ensures computational efficiency and balanced category sampling, as validated by consistent gains experiments.
	Comprehensive comparisons on the UCM, AID, and NWPU-RESISC45 datasets demonstrate that HAL outperforms classical AL baselines, and that all HAL+SSL variants (\textit{i.e.}, HSSAL) consistently exceed their SSL-only and AL-only counterparts. Particularly, the combination with FlexMatch achieves around \textbf{95\%} OA using only \textbf{6\%}, \textbf{8\%}, and \textbf{10\%} labeled samples on the three datasets, confirming the strong complementarity between hierarchical active querying and adaptive pseudo-label learning.
	
	Future work will extend HSSAL to dense prediction tasks such as semantic segmentation and change detection for enhanced robustness under domain shifts. Overall, HSSAL provides a flexible foundation for label-efficient and uncertainty-aware learning in RS applications.

	\section{Acknowledgement}
	This project is jointly supported by the German Research Foundation (DFG GZ: ZH 498/18-1; Project number: 519016653), by the German Federal Ministry for the Environment, Nature Conservation, Nuclear Safety and Consumer Protection (BMUV) based on a resolution of the German Bundestag (grant number: 67KI32002B; Acronym: \textit{EKAPEx}), and by the Munich Center for Machine Learning.
	
	\section*{Author Contributions}
	Wei Huang: Conceptualization, Methodology, Software, Validation, Formal Analysis, Investigation, Data Curation, Writing - Original Draft; Zhitong Xiong: Formal analysis, Methodology, Conceptualization, Writing, Review \& Editing; Chenying Liu: Investigation, Methodology, Formal Analysis, Writing, Review \& Editing; Xiao Xiang Zhu: Conceptualization, Methodology, Writing, Review \& Editing, Project administration, Supervision, Funding acquisition.

    \bibliographystyle{IEEEtran}
    \bibliography{IEEEabrv, mybib}

    \ifCLASSOPTIONcaptionsoff
      \newpage
    \fi

    \end{document}